\documentclass[journal]{IEEEtran}
\pdfoutput=1
\makeatletter
\def\BState{\State\hskip-\ALG@thistlm}
\makeatother

\ifCLASSINFOpdf
  \usepackage[pdftex]{graphicx}
  \usepackage{epstopdf}
  \graphicspath{{./},{./images/},{./images/challenges/}}
  \DeclareGraphicsExtensions{.pdf,.jpeg,.jpg,.png,.eps}
\else
  \usepackage[dvips]{graphicx}
  \graphicspath{{./},{./images/},{./images/challenges/}}
  \DeclareGraphicsExtensions{.eps}
\fi

\usepackage[]{algorithm2e}
\usepackage[cmex10]{amsmath}
\usepackage{array}
\usepackage[percent]{overpic}

\ifCLASSOPTIONcompsoc
  \usepackage[caption=false,font=normalsize,labelfont=sf,textfont=sf]{subfig}
\else
  \usepackage[caption=false,font=footnotesize]{subfig}
\fi

\usepackage{url}

\usepackage{color}
\usepackage{verbatim}

\newcommand{\afootnote}[2]{\textsuperscript{#1}\thanks{\textsuperscript{#1}~#2}}
\newcommand{\afootnotemark}[1]{\textsuperscript{#1}}

\begin{document}
\title{Diverse Large-Scale ITS Dataset\\ Created from Continuous Learning\\ for Real-Time Vehicle Detection}

%
\author{Justin A. Eichel\afootnote{1}{Eichel and Mishra contributed equally}, Akshaya Mishra\afootnotemark{1}, Nicholas Miller\afootnote{2}{Miller and Jankovic contributed equally}, Nicholas Jankovic\afootnotemark{2},\\%
Mohan A. Thomas, Tyler Abbott, Douglas Swanson, Joel Keller\thanks{Miovision Technologies Inc.}\thanks{Manuscript received October 15, 2014; revised December ?, 2014.}}

\markboth{IEEE TRANSACTIONS ON INTELLIGENT TRANSPORTATION SYSTEMS, VOL.~16, NO.~1, MARCH~2015}%
{IEEE TRANSACTIONS ON INTELLIGENT TRANSPORTATION SYSTEMS, VOL.~16, NO.~1, MARCH~2015}


\newcommand{\FPS}{FPS}%
\newcommand{\MB}{MB}%

\newcommand{\videoSet}{S}%
\newcommand{\videoSetPos}{{\videoSet}_{p}}%
\newcommand{\videoSetNeg}{{\videoSet}_{n}}%

\newcommand{\videoSetTraining}{{\videoSet}_{tr}}%
\newcommand{\videoSetValidation}{{\videoSet}_{val}}%
\newcommand{\videoSetTesting}{{\videoSet}_{tst}}%
\newcommand{\detectorParameters}{\Theta}%

\newcommand{\runtimeDuration}{t_{det}}%
\newcommand{\videoDuration}{t_{vid}}%

\newcommand{\I}{I}%
\renewcommand{\i}{i}%
\renewcommand{\j}{j}%

\newcommand{\Nij}{{\cal N}_{\i,\j}}%
\newcommand{\roi}{r}%

\newcommand{\m}{\vec{m}}%
\newcommand{\f}{\vec{f}}%
\newcommand{\fPos}{{\f}_p}%
\newcommand{\fNeg}{{\f}_n}%

\newcommand{\TP}{TP}%
\newcommand{\FN}{FN}%
\newcommand{\FP}{FP}%
\newcommand{\TN}{TN}%
\renewcommand{\a}{\alpha}%

\newcommand{\w}{\vec{w}}%
\renewcommand{\b}{b}%
\newcommand{\Sw}{W}%
\newcommand{\Sb}{B}%

\newcommand{\backgroundModel}{{\I}_{bg}}%

\renewcommand{\S}{S}%
\newcommand{\Sj}{{\S}_{\j}}%
\newcommand{\err}{\varepsilon}%
\newcommand{\e}{e}%
\newcommand{\D}{D}%
\newcommand{\yj}{y_{\j}}%
\newcommand{\hatyj}{\hat {\yj}}%
\newcommand{\fj}{{\f}_{\j}}%
\newcommand{\C}{C}%

\renewcommand{\c}{c}%
\newcommand{\ac}{a_{\c}}%
\newcommand{\tc}{\tau_{\c}}%
\newcommand{\h}{h}%
\newcommand{\nc}{n_{\c}}%
\newcommand{\hc}{{\h}_{\c}}%
\renewcommand{\H}{H}%
\newcommand{\bc}{b_{\c}}%
\newcommand{\dc}{d_{\c}}%
\newcommand{\fjc}{f_{\j,\dc}}%
\newcommand{\rp}{r_+}%
\newcommand{\rn}{r_-}%

\newcommand{\p}{p}%

\newcommand{\assign}[2]{[\textbf{\textcolor{red}{#1}} - \textcolor{blue}{#2}]}%
\newcommand{\comments}[1]{[\textcolor{blue}{#1}]}%
\newcommand{\needRef}[1]{[\textcolor{red}{???}]}%

\newcommand{\Tableref}[1]{Table~\ref{#1}}%
\newcommand{\Figref}[1]{Fig.~\ref{#1}}%
\newcommand{\FigPartref}[2]{Fig.~\ref{#1}(#2)}%
\newcommand{\Algref}[1]{Alg.~\ref{#1}}%
\newcommand{\Secref}[1]{Sec.~\ref{#1}}%
\newcommand{\Subsecref}[1]{Sec.~\ref{#1}}%
\newcommand{\Subsubsecref}[1]{Sec.~\ref{#1}}%

\maketitle

\begin{abstract} In traffic engineering, vehicle detectors are trained on limited datasets resulting in poor accuracy when deployed in real world applications. Annotating large-scale high quality datasets is challenging. Typically, these datasets have limited diversity; they do not reflect the real-world operating environment. There is a need for a large-scale, cloud based positive and negative mining (PNM) process and a large-scale learning and evaluation system for the application of traffic event detection. The proposed positive and negative mining  process addresses the quality of crowd sourced ground truth data through machine learning review and human feedback mechanisms. The proposed learning and evaluation system uses a distributed cloud computing framework to handle data-scaling issues associated with large numbers of samples and a high-dimensional feature space. The system is trained using AdaBoost on $1,000,000$ Haar-like features extracted from~$70,000$ annotated video frames. The trained real-time vehicle detector achieves an accuracy of at least~$95\%$ for~$1/2$ and about~$78\%$ for~$19/20$ of the time when tested on approximately $7,500,000$ video frames. At the end of 2015, the dataset is expect to have over one billion annotated video frames. \end{abstract}

\begin{IEEEkeywords}
sample selection, AdaBoost, positive mining, negative mining, real-time vehicle detection, Haar-like feature space, distributed learning and evaluation, large-scale traffic datasets.
\end{IEEEkeywords}

%
\IEEEpeerreviewmaketitle
\section{Introduction}
Automatic traffic event detection technologies play a major role in safe, reliable and efficient operations of road transportation systems~\cite{D2ITS}, including traffic surveillance~\cite{VisionSurvey1}, vehicle presence detection~\cite{DetectionSensors2}, traffic density estimation, emergency response, traffic re-routing and real-time optimized signal control~\cite{D2ITS,DetectionSensors2}. Sensing, transmitting, and computing~\cite{VANET2} are three major technological components of an automatic traffic event detector. Various sensors~\cite{MicroFerromagnetic} including road-tubes, loop-detectors, radars and cameras are used to collect measurements. The measured data are analyzed using various data analytic methods to build effective traffic management systems. Although simple non-video-based sensors can provide a higher signal-to-noise ratio than video cameras, video-based traffic measurements systems are very popular for two reasons. First, the video-based detector signal can be reviewed by humans~\cite{medina2008vol1}. Second, advanced computer vision algorithms can be employed at different stages of data collections to extract scalable information that can be used in designing efficient intelligent transportation systems (ITS)~\cite{D2ITS, Morris06,Sivaraman10}.

Many recent video-based vehicle detectors rely on machine learning to detect and classify vehicles~\cite{SivaramanLooking13}. The classifiers have a number of parameters that need to be trained to ensure that the detector correctly classifies objects of interest, such as vehicles or pedestrians, while correctly classifying `non-vehicles', such as roadway or trees. When given one or more video frames, a video-based vehicle detector might try to localize vehicles using motion, shape, and appearance-based features. For example, a Haar feature vector~\cite{ViolaJones} can be extracted locally for each region of interest in a video frame. Then, a classifier can transform the feature vector into a score, indicating how similar the region of interest is to a vehicle. The parameters controlling how the classifier transforms the feature vectors into binary or multiple class labels can be trained using samples of real traffic event data, namely images of vehicles and images of non-vehicles.

\begin{figure}
  \centering
  \includegraphics[width=0.35\textwidth]{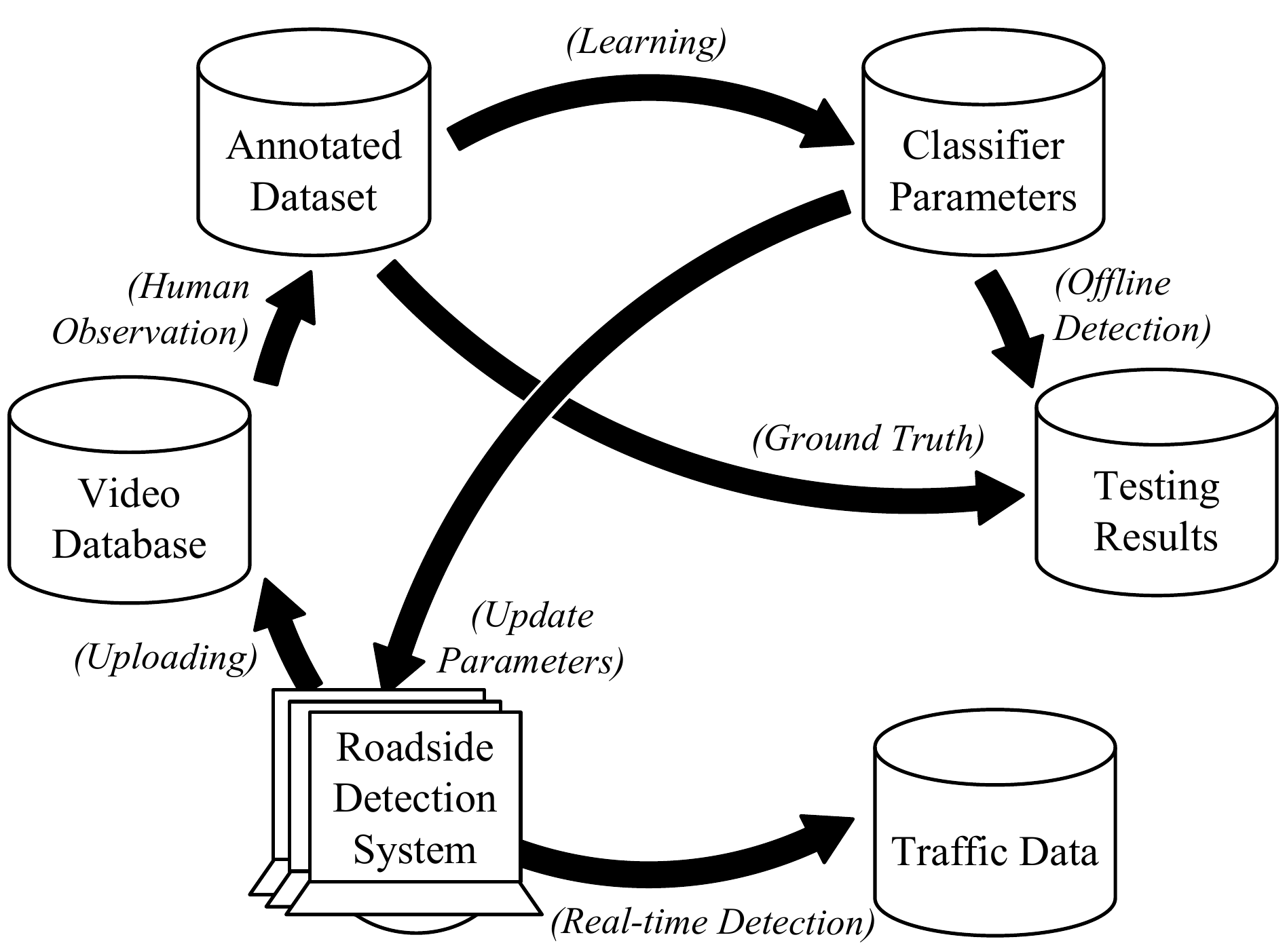}
  \caption{A scalable platform for learning and evaluating a real-time vehicle detection system. A large network of connected cameras and real-time roadside processors captures and streams traffic video. A distributed machine learning system continually samples ground truth data and supervised training examples. It learns better classification parameters and evaluates real-time vehicle detection on a large testing set. Improvements are incorporated by sending incremental parameter updates to the roadside processors. }
  \label{fig:system_summary}
\end{figure}

Many existing video detectors are trained from existing traffic datasets~\cite{saunier2014Transportation,patrick2011Learning,xiaogang2009Unsupervised,saunier2007Probabilistic,papageorgiou2000CBCL,wang2011Automatic,agarwal2004Learning,schneiderman2000Statistical,makris2001Index,bileschi2006StreetScenes}. However, many of these datasets are not actively growing, and they do not contain sufficient diversity to train, validate, and test a generalized real-time vehicle detector to be deployed in a real world application~\cite{medina2009vol4}. These datasets do not contain a sufficient number of diverse samples of weather conditions, camera perspectives, roadway conditions, and roadway configurations. Collection of this data is generally cost prohibitive due to the quantity of annotations required to sample each scenario. Maintenance of the dataset adds additional cost and it may well be infeasible to continuously add under-sampled scenarios to the dataset. Partitioning a dataset into training and test is an important consideration. Detection algorithms are most effectively trained with diverse training sets with strong representation from all decision boundaries. Recent work~\cite{NegativeMining} on weakly supervised classifier have  shown how continuous active learning using positive and negative mining can accomplish this, substantially improving the performance of general object detectors.

Further, many existing vehicle learning and evaluation systems are not designed to efficiently process billions of traffic event annotations~\cite{Sivaraman10}. Storage is required to archive the dataset, and each annotation may need to be retrieved multiple times for each training session. As new samples are continuously added to the dataset, the video detector must be periodically retrained. Distributed computing systems~\cite{sanson2012Scalable, DeepNetworks} are designed to address the storage and processing requirements. For example, Netflix stores 6.5 billion hours of video as of the first quarter of 2014 and makes extensive use of Amazon's Web Services (AWS)~\cite{AWS} for their transcoding processes. \Subsubsecref{subsec:distributed_computing_environments} provides background on AWS and other distributed computing environments upon which the proposed learning and evaluation system is built.

The authors propose three main contributions to ITS video detection: \begin{enumerate}
  \item a large-scale ITS positive and  negative data mining process for training and validation sample selection,
  \item a large-scale ITS learning and evaluation system, and
  \item a large-scale ITS traffic event dataset, soon to be available for research collaboration.
\end{enumerate} The remainder of the paper details the background, the proposed contributions, experimental results, and conclusions. The background, \Secref{sec:background}, describes the application, technology and terminology surrounding ITS vehicle detectors, (\Subsecref{subsec:vehicle_detectors}, \Subsecref{subsec:video_vehicle_detection}), the scope and availability of existing ITS related datasets, and related machine learning methods (\Subsubsecref{subsec:machine_learning_based_vehicle_detection}). \Secref{sec:proposed_solutions} presents the proposed traffic event positive and negative mining process (\Subsecref{subsec:proposed_positive_and_negative_mining}) and the proposed learning and evaluation system (\Subsecref{subsec:distributed_and_scaling_computing}), see \Figref{fig:system_summary}. The third section (\Secref{sec:experimental_validation}) presents experimental results illustrating the impact of training dataset composition on overall accuracy when evaluated on the entire testing dataset. \Secref{sec:conclusion} presents concluding remarks and discussion of future work with the intent of making the proposed ITS dataset available for research collaboration.

\section{Background}\label{sec:background}
This section provides preliminary context with a discussion of detection technology and related techniques. The section begins with a summary of vehicle detection applications and technology alternatives. Second, a broad overview of various video-based vehicle detection technologies is provided. Third, existing ITS video datasets are considered. Finally, the section concludes with a discussion of distributed computing environments and their application to the problem of training vehicle detection systems.

\subsection{Vehicle detector technology}\label{subsec:vehicle_detectors}
Road infrastructure is a significant investment made by governments and private agencies. The engineering of these traffic systems requires current usage data. Vehicle detector technologies are a key tool in collecting traffic data and extracting meaningful information from them for the purpose of building better traffic management systems. Various traffic studies, such as Automatic Traffic Recorder (ATR) studies, turning movement counts (TMCs), origin-destination (O-D) studies and travel time (TT) studies help in infrastructure budgeting, time of day traffic signal timing, vehicle density estimation for roadside advertising, toll route usage pricing calculation, and class based lane usage. Further, real-time vehicle detection technologies can help in traffic re-routing, and reduce wait times by performing demand-based signal control.

Since the introduction of inductive loops in the 1960s~\cite{TrafficDetectorHandbook}, many other intrusive sensors such as road tubes, piezoelectric cables, weigh-in-motion sensors, magnetoresistive sensors~\cite{DetectionSensors2} and micro-ferromagnetic induction coil sensors~\cite{MicroFerromagnetic} are being installed in or on the roadway to detect vehicles. Non-intrusive sensors are installed outside of the direct vehicle path, typically above the ground. Examples include microwave radar systems, infrared radar (Lidar) systems, passive infrared (PIR) sensors, ultrasonic sensors, acoustic sensors, video imaging vehicle detection systems~\cite{DetectionSensors2} and optical beam break sensors~\cite{BeamBreak}. Seismic sensors can also detect vehicles~\cite{Seismic2}; these can attach to the ground, usually beside the roadway. In addition to physical sensors, the digital era allows vehicle localization through the use of transponders, smartphones~\cite{smartphone2}, Bluetooth devices~\cite{Bluetooth} and vehicular ad-hoc networks~\cite{VANET2}.

\subsection{Overview of vehicle detection through video}\label{subsec:video_vehicle_detection}
Given all of the available technologies, real-time video offers a visual source of vehicle and environment data and does not require the vehicle or its passenger to possess any specialized technology. Video traffic data is ideally suited to learning-based computer vision algorithms~\cite{VisionSurvey2} and complements the data-driven intelligent transportation systems philosophy~\cite{D2ITS}.

In general, a vehicle detector indicates the presence of a vehicle through the following mathematical process. Measurements, $\m$, in the form of pixel intensity, are obtained for each region of interest, $\roi$, within a video frame, $\I$. To reduce complexity of the classifier, the detector takes $\m$ and extracts a feature vector  $\f$ (also referred to as ``the features''), which lies in a feature space. Ideally, given a set of annotated positive vehicle sightings $\videoSetPos$, and negative vehicle sightings $\videoSetNeg$, each feature vector $\fPos$ extracted from $\videoSetPos$ will occupy a distinct region in the feature space, differentiated from the set of feature vectors $\fNeg$ extracted from $\videoSetNeg$.

\begin{figure}
  \centering
  \newcolumntype{M}{>{\centering\arraybackslash}m{\dimexpr.45\linewidth-2\tabcolsep}}
  \begin{tabular}{MM}
    \includegraphics[width=4cm,height=2cm]{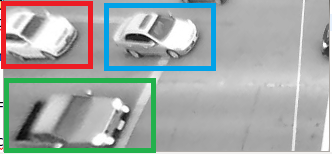} & \includegraphics[width=4cm,height=2cm]{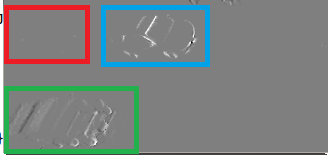} \\
    (a) source: close up & (b) motion: close up \\
    \\
    \includegraphics[width=4cm,height=3cm]{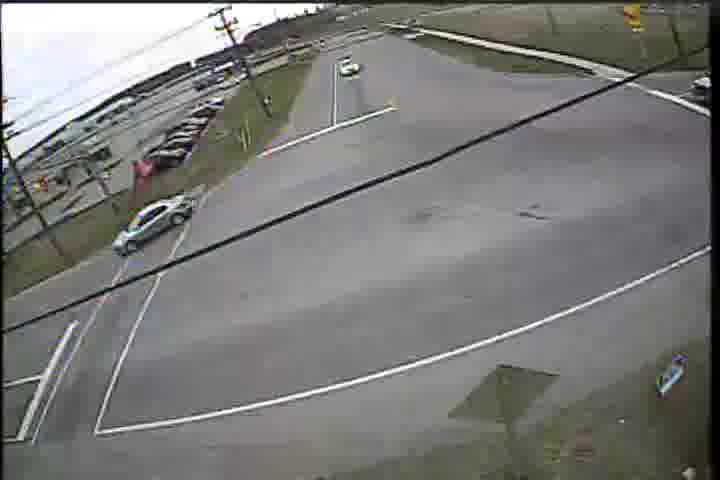} & \includegraphics[width=4cm,height=3cm,trim=15mm 15mm 15mm 15mm,clip]{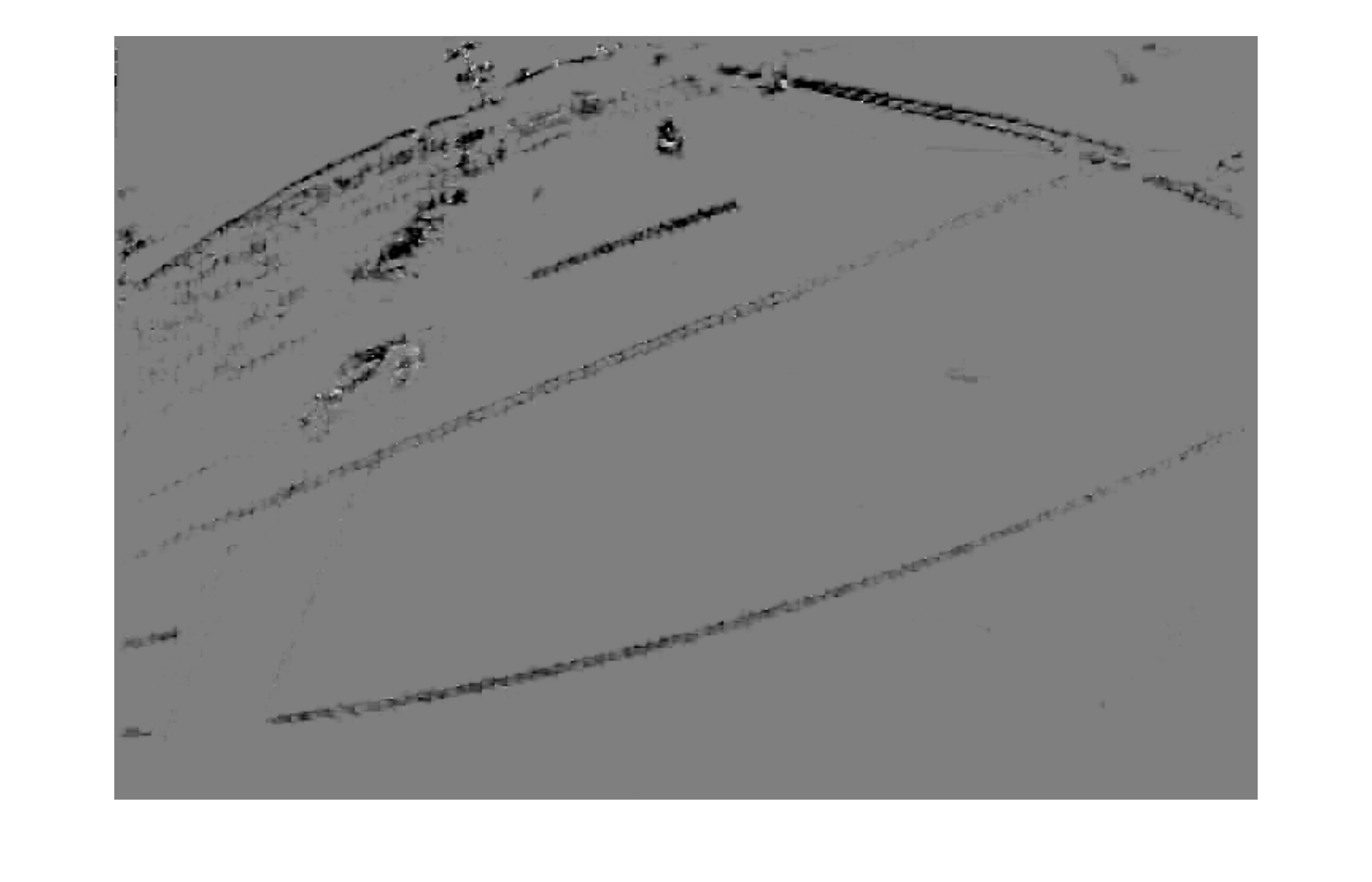} \\
    (c) source: windy & (d) motion: windy
  \end{tabular}
  \caption{Given (a) a roadway, Harrison's implementation of a Reichardt motion model~\cite{harrison2000Robust,vanSanten1985Elaborated} creates (b) strong responses for the moving vehicles, (red) and (blue), but fails to create any response for the stationary vehicle (green). Given  (c) a different roadway location containing minor wind conditions, the motion response (d) is strong for moving background scenery, making it difficult to determine which motion responses correspond to background and which ones correspond to vehicles.}
  \label{fig:failure_of_motion_only}
\end{figure}

Existing video-based vehicle detector feature spaces typically fit into two categories, (a) motion-based and (b) appearance and geometry-based features. Motion-based features allow the detector to identify moving vehicles from temporal changes over a set of consecutive video frames. Motion-based video detector algorithms are intuitive, easy to implement, computationally efficient, and can be implemented in real-time using low cost embedded systems. However, if $\backgroundModel$ does not account for dynamic changes due to illumination, glare, rain, snow, fog, wind, or other weather-related effects, the detector may fail to distinguish between dynamic background regions and moving vehicles~\cite{medina2008vol1, medina2009vol4}. Further, the performance of motion-based detectors degrades significantly when the object of interest is stationary or moves slowly. The detector may also misclassify vehicles in the presence of stationary or independently moving motion fields, such as if the vehicle is occluded due to trees or overhead wires. As exemplified in \Figref{fig:failure_of_motion_only}, vehicle detectors using only motion features cannot accommodate the aforementioned real-world scenarios~\cite{medina2009vol4}. On the other hand, machine learning based vehicle detection using shape features has shown promising results for classifying internet images~\cite{DeepNet1}.

\subsection{Machine learning based vehicle detection}\label{subsec:machine_learning_based_vehicle_detection}
\begin{figure}
    \centering
    \newcolumntype{M}{>{\centering\arraybackslash}m{\dimexpr.32\linewidth-2\tabcolsep}}
    \begin{tabular}{MMM}
      \includegraphics[height=2.75cm,width=2.75cm,trim=0mm 2mm 0mm 2mm,clip]{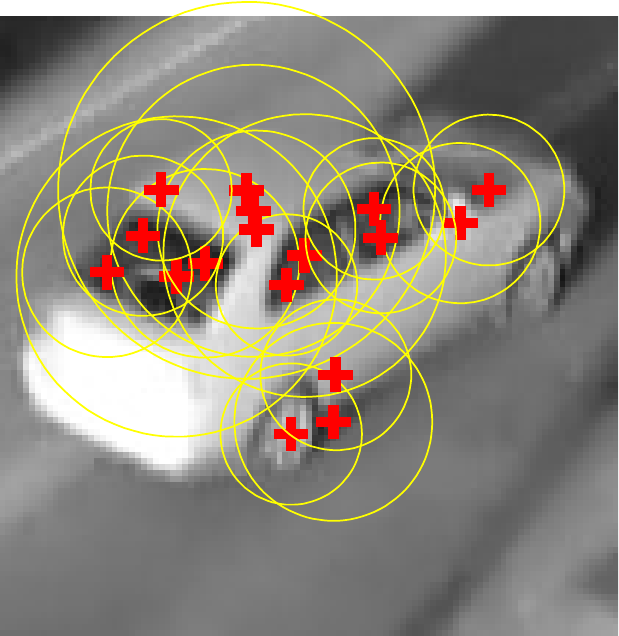} & \includegraphics[height=2.75cm,width=2.75cm,trim=0mm 2mm 0mm 2mm,clip]{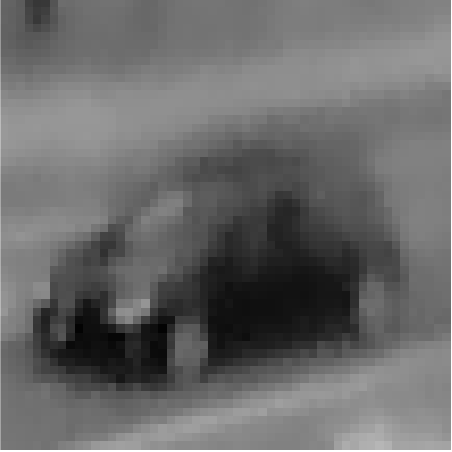} & \includegraphics[height=2.75cm,width=2.75cm,trim=0mm 2mm 0mm 2mm,clip]{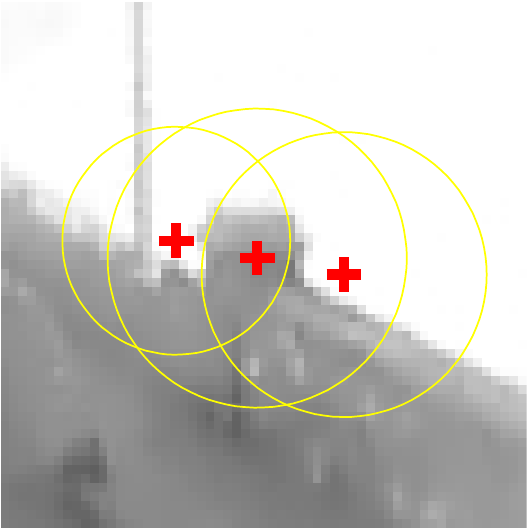} \\
      (a) correct & (b) missing & (c) erroneous
    \end{tabular}
    \caption{SURF keypoints~\cite{SURF} detected using constant sensitivity parameters. Note that the car in (a) has keypoints which may be matched for detection, while the car in (b) has no SURF keypoints. Furthermore, a background object, a bridge, in (c) has keypoints, which could result in a false positive detection.}
    \label{fig:failure_of_sift}
\end{figure}

The three main components of machine learning based detectors are feature extraction, feature selection and classifier design. Typically, feature extraction techniques estimate salient features, $\f$, based on a vehicle's appearance and shape. Such features can be calculated at various spatial scales, locally at a single pixel location, $\bigl( \i, \j \bigr)$, in $\I$, regionally for a neighborhood, $\Nij$, surrounding $\bigl( \i, \j \bigr)$, or globally over all of $\I$. Many appearance and geometry-based features include Haar-like features~\cite{ViolaJones}, histogram of oriented gradients (HOG)~\cite{HOG}, scale-invariant feature transform (SIFT)~\cite{SIFT}, SURF~\cite{SURF}, ORB~\cite{ORB}, Gabor filters~\cite{Gabor}, and super pixels~\cite{Krig2014}. The selection of appropriate features varies by application depending on the properties of the class that will be detected. For example, ~\Figref{fig:failure_of_sift} illustrates  the SURF feature response applied to real-world ITS video produces strong responses for both vehicles and non-vehicles, while a classifier trained on Haar-like features~\cite{MiovisionCVPR2014}, produces strong responses for vehicles and weaker responses for non-vehicles. The proposed learning and evaluation system in~\Secref{sec:proposed_solutions} is based on Haar-like features applied at various spatial scales.

Further, since each feature has a computational and memory cost and video detectors must operate on cost effective hardware in real-time, the process of feature selection is useful to determine which subset of features, from the set of all possible features, contribute the most to the video detector accuracy. Fortunately, the feature selection process can be performed offline using principal components analysis (PCA), independent component analysis (ICA)~\cite{FanICA07}, and unsupervised clustering techniques~\cite{Morris06}. Other feature selection techniques are built into classifier training. For example, support vector machines (SVM), bagging and boosting~\cite{Freund1997119}, and convolutional neural networks~\cite{DeepNet1}, recurrent neural networks and incremental recurrent neural networks~\cite{DeepNetworks} determine which features are significant as part of their training process. Computational performance is improved since only the selected features are calculated as the real-time vehicle detector evaluates each $\roi$ in each $\I$.

Once features are defined, the classifier is then trained such that a given $\f$ is classified as either a vehicle or non-vehicle based on the how similar $\f$ is to the collection of all $\fPos$ or the collection of all $\fNeg$. A simple classifier, such as nearest-neighbour~\cite{Morris06}, may assign $\f$ as a vehicle if the distance of $\f$ to the nearest example in $\fPos$ is less than the distance of $\f$ to the nearest $\fNeg$ sample. Based upon applications, classifier complexity can increase to produce advanced classifiers such as neural networks (NN), cascading classifiers (CC), boosted classifiers, and support vector machines (SVM). Boosting methods are derived from the idea that many simple classifiers can be combined to be more accurate than any one simple classifier. Freund et al.~\cite{friedman2000} detail how a collection of discriminants, each with a classification accuracy of at least $50\%$, can be combined into a single classifier with significantly higher classification accuracy~\cite{Freund1997119, Sivaraman13}.

The real-time and real-world classification performance of  machine learning based vehicle detectors significantly depends on the availability of a large-scale high quality annotated vehicle dataset and a scalable learning and evaluation system. The scale and scope of several major ITS datasets published between 1998 and 2014 are summarized in \Tableref{tab:its_datasets}. The creation of a large high quality dataset requires a good positive and negative mining framework as well as dedicated resources to perform the review tasks~\cite{LongServedio2008}.

\subsubsection{Distributed computing environments}\label{subsec:distributed_computing_environments}
Given a large amount of data and a high dimensional feature space, one method of processing the data is to construct high performance computing (HPC) infrastructure through computer clusters~\cite{moore2011Trestles}. These cluster infrastructures provide methods for sharing memory and distributing computations over a large number of computers. Although the process of building and configuring a cluster has been simplified~\cite{papadopoulos2004Configuring}, this solution requires capital investment and maintenance costs, which requires in-house technology experts. Further, the process of designing an algorithm and evaluating it on the dataset requires almost no demand during the design process, but high computational demand during the evaluation process; demand for HPC may be high, but variable, and the cluster has insufficient capacity to complete jobs in a timely manner. Elastic cloud computing offers an alternative option which can dynamically scale to match variable work loads, such as on-demand video transcoding~\cite{jokhio2013Prediction}. For instance, Amazon Web Services (AWS)~\cite{murty2008Programming} offers distributed memory storage connected through high speed networks to on-demand computing systems. Human annotators can generate data from anywhere in the world and the learning and evaluation system can scale to accommodate large-scale datasets of annotated video. The authors propose a large-scale learning and evaluation system, built on top of AWS, that is capable of training and evaluating a vehicle traffic detector on billions of annotations.

\begin{table*}
    \centering
    \caption{ITS Datasets}
    \label{tab:its_datasets}
    \begin{tabular}{|ll|rrr|ll|l|l|}
    \hline
    Year & Author & Num. Videos & Num. Annotations & Resolution & Color & Classes & Description \\
    \hline
    2000 & Schneiderman~\cite{schneiderman2000Statistical}    & 213   & 213   & various & gray & car & various still images\\
    2000 & Papageorgiou~\cite{papageorgiou2000CBCL}           & 516   & 516   & $128 \times 128$ & color  & car   & still images with head-on view \\
    2001 & Makris~\cite{makris2001Index}                      & 20    & 0     & $640 \times 480$ & color  & none   & various parking lot views \\
    2004 & Agarwal~\cite{agarwal2004Learning}                 & 1328  & 1,050 & $100 \times 40$ & gray & car & 278 testing images at various scale \\
    2006 & Bileschi~\cite{bileschi2006StreetScenes}           & 3547  & 27,666 & $1280 \times 960$ & color & many & still images of cars, pedestrians and more \\
    2007 & Saunier~\cite{saunier2007Probabilistic}            & N/A   & 2941  & N/A & gray & N/A & 10 to 60-sec clips from common location \\
    2007 & Saunier~\cite{saunier2007Probabilistic}            & 1     & 47,084 & N/A & N/A & N/A & 1-hour segment at an intersection \\
    2009 & Kasturi~\cite{kasturi2009Framework}             & 100   & $\sim37,500$ & $720 \times 480$ & color & car & $\sim2.5$ min
    videos with annotated I-frames \\
    2009 & Xiaogang~\cite{xiaogang2009Unsupervised}           & 1     & 540   & $720 \times 480$ & color & car,ped & 1.5-hour segment at an intersection \\
    2011 & Patrick~\cite{patrick2011Learning}                 & 630   & 315   & $216 \times 384$ & gray   & car   & optical flow data for each frame \\
    2011 & Wang~\cite{wang2011Automatic}                      & 1     & 520   & $720 \times 480$ & color & ped & 90-min far-field intersection\\
    2014 & Saunier~\cite{saunier2014Transportation}           & N/A & N/A     & $640 \times 480$ & color & car,ped &  1-intersection, 4-cameras, 3-months \\
    2014 & Saunier~\cite{saunier2014Transportation}           & N/A & $\sim1,000$ & $800 \times 600$ & color & car,ped &
    2-intersection, 1-cameras, 2-hours each  \\
    \hline
    \textbf{2014} & \textbf{Proposed} & 1,718 & \textbf{7,731,000} & various & color & car & 5-min video segments \\
    & \textbf{Miovision} & & & & & & 19,244 training samples \\
    & & & & & & & time-of-day, weather \\
    & & & & & & & various road conditions \\
    & & & & & & & 15 locations \\
    \hline
    \end{tabular}
\end{table*}

\subsection{Active and continuous learning}
During the training phase, it is not uncommon \cite{kasturi2009Framework, Morris06} to use manually partitioned datasets to test and train with some number of positive and negative samples. Tamersoy et al.~\cite{Tamersoy2009} used `difficult' negative samples, containing vehicle components, with the intention of training more robust detection algorithms. Such an approach is defensive; the training set is selected in anticipation of likely failure modes in the detection algorithm. Including `difficult' samples in the training set can improve classifiers at otherwise ambiguous decision boundaries.

Sivaraman~\cite{Sivaraman10} demonstrated how an active learning approach can effectively be used to train a robust real-time on-road vehicle tracking algorithm. In their framework, two training iterations were used to focus on informative samples. The first iteration trains the detector with a manually partitioned data set. The trained detector is then evaluated with an independent test set. Results of this evaluation are selectively sampled to augment and prune the original training set in such a way that difficult decision boundaries are more heavily represented. The effect of this was to create a more robust detector, reducing the number of false positives. Positive and negative mining techniques are an important component of active and continuous learning because they reduce a large pool of datasets into a manageable and representative set~\cite{NegativeMining}. A classifier trained on samples selected using positive and negative mining provides better classification accuracy compared to a classifier trained on general samples~\cite{NegativeMining}.~\Tableref{tab:its_datasets} includes several datasets appropriate for training, and in some cases, testing video detection algorithms.

\subsection{Summary of issues related to ITS vehicle detection systems}
\subsubsection{ITS Datasets}
Existing datasets such as \cite{schneiderman2000Statistical} and \cite{kasturi2009Framework} include annotations to localize general vehicle objects, yet do not provide detail about scene conditions and operating environments. Although there are a few domain specific annotations such as vehicle class present, they are limited in scope. They may be limited in number of locations or perspectives. Very often, diverse weather conditions and lighting conditions are not present, and where they are, these conditions are not annotated. The datasets did not annotate roadways or intersections and, consequently, could not provide lane geometries or lane assignment of vehicles. Generally, training a detector against datasets with such limited scope does not provide confidence for real-world performance. Further, there is no information regarding the data acquisition and annotation process - a notable exception is Kasturi et al~\cite{kasturi2009Framework}, which demonstrated the benefits of a formal annotation review process. However, in general, basic questions including,   ``are the videos coming from different sources?'',  ``was the annotation process audited for reliability?''~\cite{LongServedio2008}, and ``what are the conditions represented?'' remain unanswered with most of the datasets.

\subsubsection{Learning and evaluation platform}
Carefully designed sample sets and feature vectors play a significant role in the success of machine learning based vehicle detection system~\cite{LongServedio2008}. Selecting an optimal set of training samples and features vectors that produce minimal generalization error, require a daunting amount work in designing a scalable computational frame work. Existing learning and evaluation platforms are unable to handle billions of samples and millions of features from which an optimal set of quality training samples and feature vectors can be selected. Continuous positive and negative mining tools have been effectively used for general purpose computer vision detection and tracking applications~\cite{RothActiveSampling}, but they have not yet been applied to large-scale ITS datasets.

\section{Proposed solutions}\label{sec:proposed_solutions}
As discussed in \Subsecref{subsec:machine_learning_based_vehicle_detection}, to the best of the authors' knowledge, a comprehensive ITS traffic dataset for vehicle detection does not exist; existing datasets do not contain sufficient diversity to train a vehicle detector for deployment in real-world conditions. As a result, the fundamental objective of this paper is to develop a large-scale diverse traffic dataset, through a data mining process for semi-supervision of learning algorithms, and a learning and evaluation platform, for estimating the optimal parameters of a given classifier and feature extraction method.

\begin{figure}
    \centering
    \includegraphics[width=0.35\textwidth]{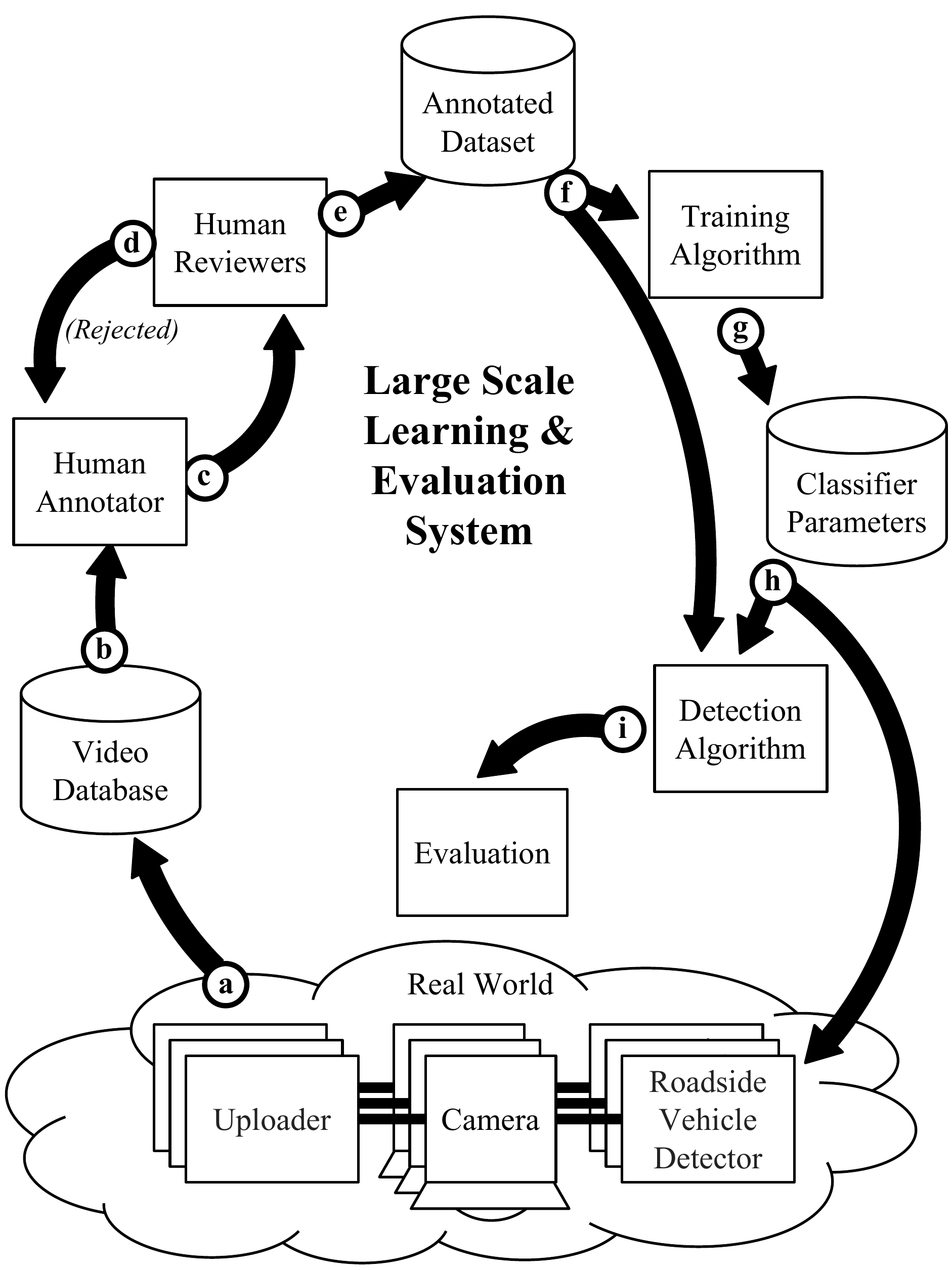}
    \caption{Overview of the Large-Scale Machine Learning System: A network of cameras, each with a roadside processor, are deployed at signalized intersections. The processor records traffic video segments, which can be uploaded to the Video Database (a,b). Random still frames are sampled from the database and assigned to Human Image Observers (c), which annotate the frame to assign ground truth vehicle observations. All annotated observations are stored within the Training Samples Database (d). These samples are then used by the Training Algorithm (e) to generate Classifier Parameters (f). From the video segments (g) and the learned parameters (h) the Detection Algorithm generates vehicle observations (i). The same video segments are also annotated by a Human Video Observer to generate ground truth vehicle observations. The Evaluation step compares the detected (i) and annotated (j) observations to measure the effectiveness of the newly trained Detection Algorithm.}
    \label{fig:learning_and_evaluation_overview_process_diagram}
\end{figure}

\Figref{fig:learning_and_evaluation_overview_process_diagram} illustrates the composition of a large-scale learning and evaluation system. The system contains data sensing, through real-world cameras, and a human annotation process that labels objects of interest in video frames. A training algorithm, based on dataset mining, is described in \Subsecref{subsec:proposed_positive_and_negative_mining}, which utilizes a parallelizable AdaBoost classifier described in \Subsecref{subsec:classifier_parameter_estimation}. A large-scale distributed computing system, described in \Subsecref{subsec:distributed_and_scaling_computing}, is used to host the training and evaluation algorithms, allowing for efficient and parallelized processing. The resulting classification parameters can be distributed to live production systems once validated using the evaluation component.

\subsection{Problem formulation}\label{subsec:problem_formulation}
First, a general classifier, $\C$ is presented. $\C$ operates on a labeled sample, $\Sj$, with known class correspondence, $\yj$, and with a corresponding feature vector, $\fj$. Let $\D$ represent the entire population of traffic events, and $\S$ be the complete set of samples, a subset of $\D$, used in training and validation. The classifier generates labels $\hatyj$ such that \begin{align}
   \hatyj &= \C \bigl(\fj \bigr).
\end{align} The ideal classifier minimizes the generalization error, $\err$, the difference between each $\hatyj$ and the actual value, $\yj$, \begin{align}
  \err &= \sum_{\j \in \D} \begin{cases}
    0,	& \text{if~} \hatyj = \yj \\
    1, 	& \text{else}.
  \end{cases}
  \label{eq:general_error}
\end{align} In practice, it is not possible to sample the entire population as implicitly indicated in~\eqref{eq:general_error}. Only a sampled population, $\S$, is available. $\S$ must be representative of $\D$: a representative sample population is the cornerstone of positive and negative mining.

The parameters of $\C$ can then be estimated by minimizing $\err$ using $\S$ instead of $\D$. The discriminant function $\C$ can be represented using a simple complex function, or a set of weak functions or classifiers. In this paper, $\C$ is represented as discriminant function using a set of weak classifiers $\h$, where each weak classifier $\hc$, contains a slope $\ac$, an offset $\bc$, and a threshold $\tc$ that segments a specified feature dimension $\dc$, into two regions, $\rp$ and $\rn$~\cite{RegStump}, where \begin{align}
  \rp &= \begin{cases}
    1,	& \text{if~} \hc \geq 0 \\
    0, 	& \text{else}.
  \end{cases} &, & &
  \rn &= \begin{cases}
    1,	& \text{if~} \hc < 0 \\
    0, 	& \text{else},
  \end{cases}
\end{align} and \begin{align}
  \hc = \ac \H\bigl(\fjc - \tc \bigr) + \bc~\cite{RegStump},
\end{align} using $\H$ to represent the Heaviside step function. The complexity of the classifier depends on the number of weak classifiers, $\nc$, which should also be minimized for computational efficiency. Using all of $\hc$, $\C$ is defined \begin{align}
  \C = \text{sign}\Biggl( \sum_{c=1}^{\nc} \hc \Biggr).
\end{align}

Given this formulation, there are two required steps in order to train a generalized real-time vehicle detector. First, a sampling process~(\Subsecref{subsec:proposed_positive_and_negative_mining}) must be established to obtain a representative $\S$ from $\D$ efficiently. Second, the classifier parameters, $\nc$ and $\{\ac, \bc, \tc, \dc\} \forall \c$, must be estimated using a distributed computing system~(\Subsecref{subsec:distributed_and_scaling_computing}).

\subsection{Positive and negative mining}\label{subsec:proposed_positive_and_negative_mining}
\begin{figure}
    \centering
    \includegraphics[width=0.35\textwidth]{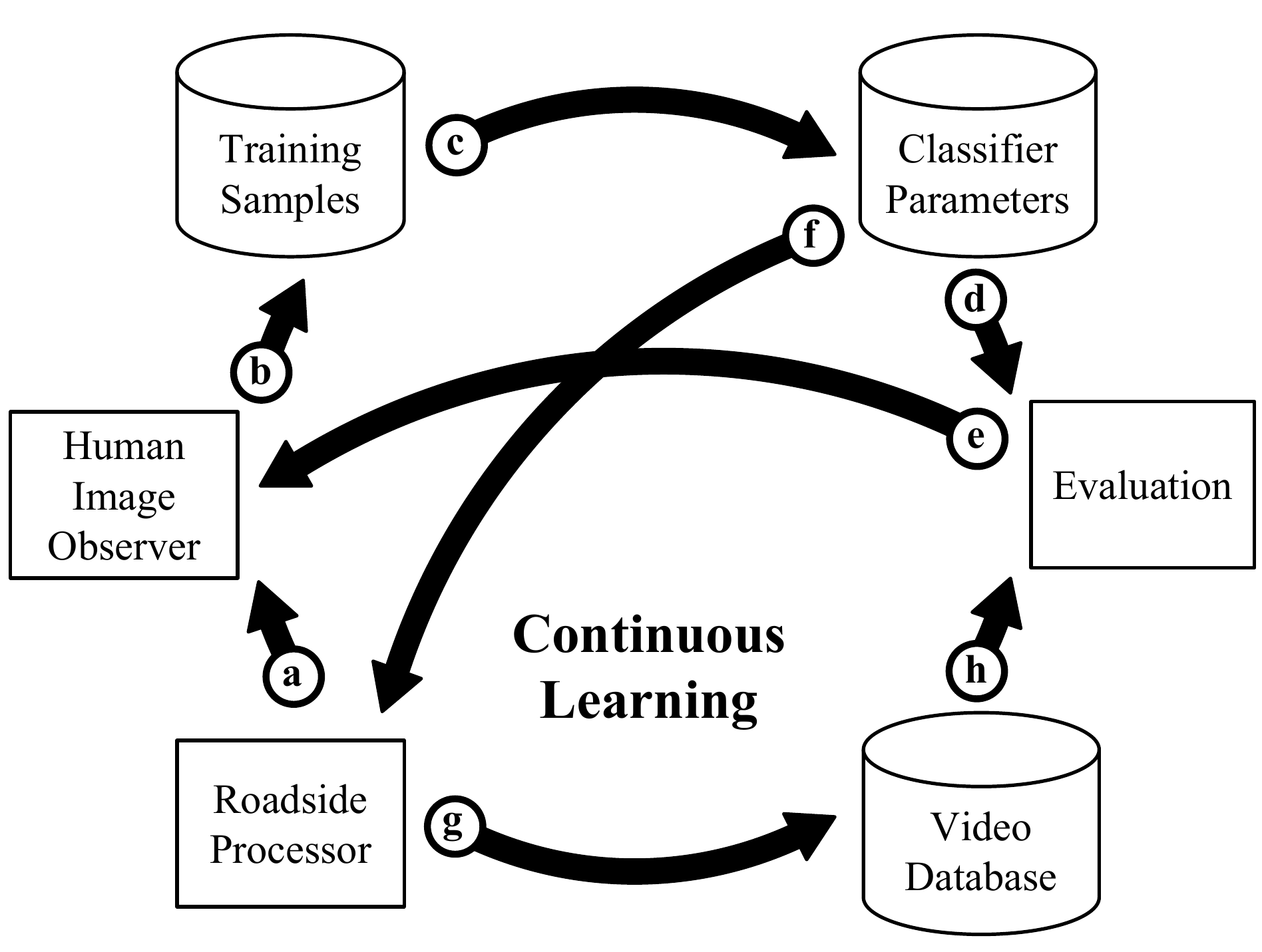}
    \caption{Continuous Learning is a simple extension to the system whereby the Roadside Processor executes the current Detection Algorithm and randomly samples the video stream, indicating the presence or absence of vehicles. These samples are uploaded and assigned to a Human Image Observer (a), who independently annotates the ground truth and adds it to the Training Sample Database (b). Updated samples are used during Training (c) to produce new Classifier Parameters, which generate updated vehicle observations (d). Falsely classified observations are fed back to a Human Image Observer (e), who adds them to the Training Sample database. Should the new Classifier Parameters show improvements, they can be downloaded to the Roadside Processor in the field (f). Although much more costly, it is still possible to upload new video segments from the Roadside Processor (g) to continuously grow the testing set for Evaluation (h).}
    \label{fig:continuous_learning}
\end{figure}

Data mining is required to select a representative $\S$ from $\D$. The major steps are outlined in~\Algref{alg:pos_neg_mining}, and the data model is illustrated in~\Figref{fig:continuous_learning}. First, the mining begins with an initial manually dataset, $\S$, of carefully chosen positive and negative samples from $\D$. Then, the estimated parameters for $\C$, that best classify vehicles and non-vehicles, are determined using $\S$. The classifier is evaluated on a percentage, $\p$, of random samples from $\D$, and are compared to corresponding human annotations. Misclassified samples are then added to $\S$ and, using bias and variance analysis~\cite{Freund1997119}, noisy and overrepresented samples contained in $\S$ are removed. The process is repeated until the $\err$ achieves a minimum, the $\C$ complexity, $\nc$, exceeds a threshold, or a maximum number of iterations is achieved.

\begin{algorithm}
 \KwIn{$\S$ manually curated initial sample of $\D$ with corresponding $\yj$}
 \KwOut{$\S$ containing a representative sample of $\D$}
 $\i \gets 0$\;
 \While{isConverging$(\err, \C, \i)$}{
  $\C \gets$ estimateClassifierParameters$(\S, \{\ldots, \yj, \ldots\})$\;
  $\S^{*} \gets $ sparseRandomSampling$(\D, \p)$\;
  \tcc{evaluate sample labels}
  \ForAll{$\j$}{
    $\yj^{*} \gets $ getManualLabel$\bigl(\Sj{*}\bigr)$\;
    $\hatyj^{*} \gets \C\bigl( \Sj^{*} \bigr)$\;
  }
  $\err \gets $calculateError$(\hatyj^{*}, \yj^{*})$\;
  \tcc{add misclassified samples to dataset}
  \ForEach{$\hatyj^{*} \neq \yj^{*}$}{
    $\S \gets \S \cup \Sj^{*}$\;
  }
  $\S \gets $rejectionSampling$(\S)$\;
  $\i \gets \i+1$\;
 }
 \caption{Positive and negative mining}
 \label{alg:pos_neg_mining}
\end{algorithm}

The data mining algorithm is the training algorithm represented in~\FigPartref{fig:learning_and_evaluation_overview_process_diagram}{f-g}. The function, estimateClassifierParameters(S), is implemented using AdaBoost, described in~\Subsecref{subsec:classifier_parameter_estimation} and~\Figref{fig:learning_flow}.

\subsection{Classifier parameter estimation}\label{subsec:classifier_parameter_estimation}
Adaptive Boosting, or AdaBoost, is a well-known algorithm for efficiently building a single strong classifier, $\C$, from a collection of weak classifiers, $\h$. At each iteration, AdaBoost attempts to estimate $\{\ac, \bc, \tc, \dc\}$ for a single weak classifier, $\hc$, in this case one dimensional regression stumps, one for each feature dimension, that can best segment vehicles, $\fPos$, from non-vehicles, $\fNeg$, with the minimal weighted classification error. Although initially each sample contributes equally when calculating classification error, samples with the greatest classification error are given more weight when computing error during subsequent iterations while samples with the least error are given less weight. Details of an AdaBoost implementation are provided by Friedman~\cite{friedman2000}. All parameters of $\C$ are determined once the algorithm converges on a minimal $\err$ or maximum $\nc$. In the past, Sivaraman et al. implemented AdaBoost for a driver assistance program in 2013~\cite{Sivaraman13}. The pseudo-code for a AdaBoost algorithm  is shown in~\Algref{alg:adaboost}.

\begin{algorithm}
  \KwIn{$(\S, \vec{y})$}
  \KwOut{$\h$}
  $\h \gets \{\}$, $nj \gets \sum_{\forall \j} 1$, $\c \gets 0$\;
  $w_{\j,0} = \frac{1}{nj} \forall \j$\;
  $\fj \gets $extractFeatures$(\Sj) \forall \j$\;
  $n_f \gets $numFeatureDimensions$()$\;
  $\e \gets 0 \cdotp [1 \ldots n_f]$\;
  \While{isConverging($\c, \e)$}{
    \tcc{update weights}
    \ForAll{\j}{
      $w_{\j,\c} \gets \exp\biggl(-\yj \C\bigl(\fj\bigr) \bigr)$
    }
    \tcc{calc candidate classifiers}
    $\e \gets 0$\;
    \For{ $\i=1:n_f$}{
	$\dc \gets \i$\;
	\tcc{adaboostEst implements~\cite{friedman2000, RegStump}}
	$\{\ac, \bc, \tc, \dc, \e_{\i}, w^\i\} = $adaboostEst$(\f, \vec{y}, w, \c)$\;
	${\hc}_{\i} \gets \{\ac, \bc, \tc, \dc, \e_{\i}\}$\;
    }
    $w \gets w^{\i^*}$, $\hc \gets {\hc}_{\i^*},$ such that $\i^* = \arg \min_{\i} \e_{\i}$\;
    $\h \gets \h \cup \hc$\;
    $\c \gets \c+1$\;
  }
 \caption{AdaBoost iteration}
 \label{alg:adaboost}
\end{algorithm}

\begin{figure}
   \centering
  \newcolumntype{M}{>{\centering\arraybackslash}m{\dimexpr.45\linewidth-2\tabcolsep}}
  \begin{tabular}{MM}
    \includegraphics[width=3cm,height=2cm]{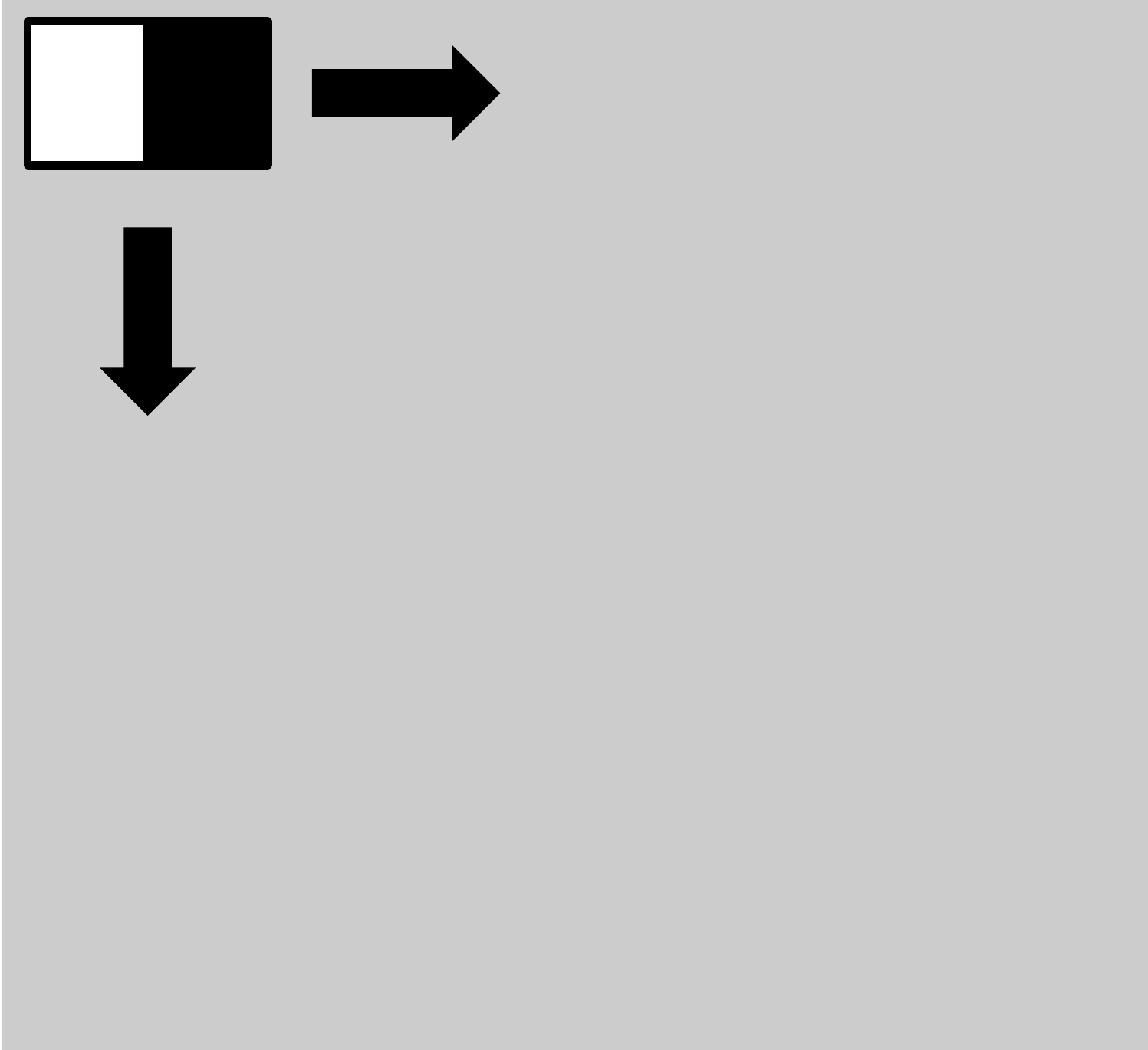} & \includegraphics[width=3cm,height=2cm]{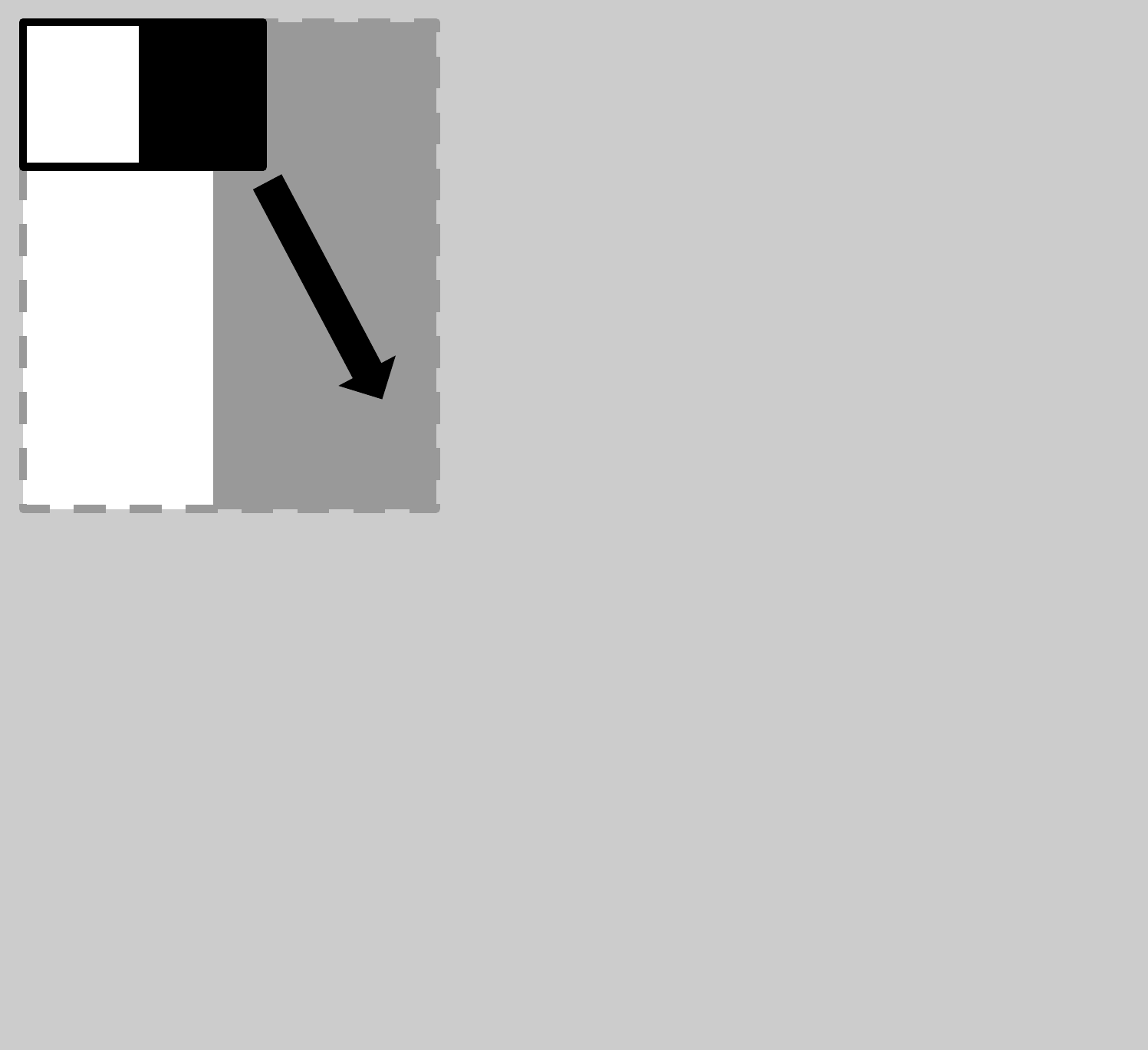} \\
    (a) Translations & (b) Scales \\
  \end{tabular}
  \caption{An example Haar kernel on an image patch (gray). There is a feature value associated with (a) all translations and (b) all scales for each translation. }
  \label{fig:haar_dimensionality}
\end{figure}

Features, $\fj$, used in this implementation are computed using the conventional Haar-like kernel. Each feature is calculated by multiplying a kernel, containing ones and negative ones, with pixel intensities from an image patch. However, when traversing all scales and translations of each Haar-kernel,the resulting feature-dimensionality becomes large. An $n_x \times n_y$ resolution image patch contains $n_t$ possible kernel translations and $n_s$ possible kernel scales, where $n_t$ and $n_s$ are both equal to $n_x \times n_y$; the kernel can be centered at any pixel and the kernel can have an area ranging from $1$ to $n_x \times n_y$ pixels squared, see~\Figref{fig:haar_dimensionality}. The total number of possible features, $n_f$, for $n_k$ potential kernels, is \begin{align}
  n_f &= n_t n_s n_k = \bigl(n_x n_y\bigr) \bigl(n_x n_y\bigr) n_k = {n_x}^2 {n_y}^2 n_k.
\end{align} For a $42 \times 42$ resolution image patch with eight unique kernels, $n_f = 24,893,568$, neglecting boundary conditions for simplification.

\begin{figure}
    \centering
    \includegraphics[width=0.38\textwidth]{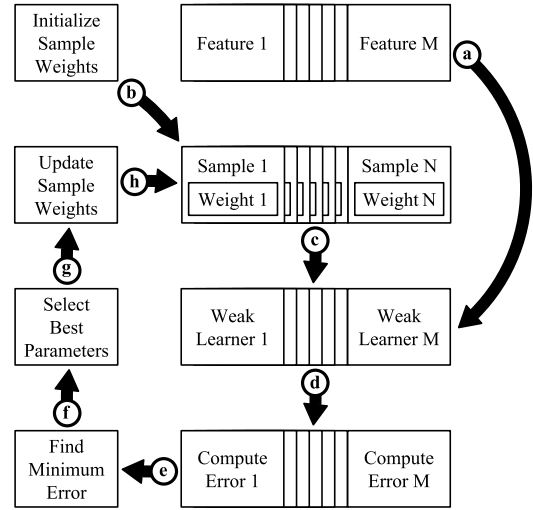}
    \caption{First, for this distributed learning system, the Haar like features are extracted and sent to an iterative AdaBoost learning system, where multiple processes compute weak learners corresponding to a single feature.  Then a master machine combines all the learner to select a best learner. Further, the best parameters are used to update the sample weights for the next iteration.}
    \label{fig:learning_flow}
\end{figure}

To account for such a high-dimensional feature space, the proposed learning system, shown in~\Figref{fig:learning_flow}, integrates a parallelized AdaBoost implementation to estimate an optimal set of parameters. The following AdaBoost components, \begin{enumerate}
  \item Haar-like features, $\fj$,
  \item sample weights, $w$,
  \item and weak-classifier candidates, ${\hc}_{\i}$
\end{enumerate} can be calculated independently and are processed in parallel. A distributed computing system can implement these steps through parallel processing and distributed memory.

\subsection{Distributed and scalable computing}\label{subsec:distributed_and_scaling_computing}
Due to the iterative nature of algorithm design, the training and evaluation system executes many times during the course of development as parameters are tuned and as new algorithm ideas are integrated. As researchers propose changes, feedback is necessary to determine if their changes improve accuracy on desired real-world scenarios. Evaluating the dataset in a reasonable amount of time is critical to a practical development process. This section outlines the requirements, workflow, and tools for a distributed computing environment capable of efficient processing.

The evaluation system must be robust to errors occurring during an evaluation; the system must provide feedback and fail gracefully, especially if costs are incurred if the evaluation system continuous to produce erroneous results. Consecutive runs of the same algorithm on the same dataset should be deterministic and produce the same results and recover from unexpected network issues or individual processor failures. The evaluation system must also be easy to use so that researchers can focus on video detectors and do not need to worry about the underlying evaluation infrastructure. The system must be able to accommodate additional data added or removed from the annotation dataset, and must allow evaluation on subsets of the dataset. The accuracy and performance results of the candidate vehicle detection algorithm must then be reported and stored for future reference.

The distributed system workflow is an implementation of the general MapReduce pattern~\cite{ekanayake2008MapReduce}. \begin{enumerate}
  \item Select the complete or a query defining a subset of the annotated dataset.
  \item Upload or specify a vehicle detection algorithm.
  \item Select an appropriate set of trained parameters.
  \item An evaluation task is created for each localization annotation from the dataset specified above.
  \item The evaluation tasks are dispatched to processors.
  \item Individual processors obtain required data and execute the evaluation task.
  \item Accuracy and performance results are aggregated.
  \item Statistics are calculated from the aggregation.
  \item Results are reported and achieved for future comparison.
\end{enumerate} The task creation, dispatch, processing, and aggregation steps are executed on numerous Amazon EC2~\cite{murty2008Programming} instances using Amazon S3~\cite{murty2008Programming} for annotation and video data storage. Large amounts of video can be shared between EC2 instances and S3 using internal Amazon high speed networks. By default the number of EC2 instances is limited to $20$ simultaneous instances per instance type, but can be increased if needed, and each instance currently contains up to $32$ cores. The EC2 solution has sufficient computing power for the current dataset needs.

\section{Experimental validation}\label{sec:experimental_validation}

\subsection{Data acquisition}\label{subsec:data_acquisition}
Data collected for this study is obtained from prototype Miovision Permanent Connected Intersection Count Stations, distributed over $15$ locations. While GIS locations are recorded for each intersection, such information is anonymized, but access to time of day information and a broad provincial context is available to researchers. Stations operate at the roadside in all environments, day and night. Videos encompass the entire intersection because stations are equipped with fisheye lenses. The output video is encoded to H.264 with a resolution of $1536 \times 1536$ at $15$ frames per second. The processor then rectifies regions of video into perspective views that contain sets of adjacent lanes.

In 2014, data obtained from the portable Miovision Scout Video Collection
Units~\cite{mcbride2012method} consist of approximately $52,000$ unique North American locations, intersections and roadways, with an additional $9,000$ across Europe. A subset of this data is currently being added to the proposed traffic event dataset and the authors intend to integrate portions of 2015 data as it is collected.

\subsection{Video dataset}\label{subsec:annotation_process}
\begin{figure*}
    \centering
    \newcolumntype{M}{>{\centering\arraybackslash}m{\dimexpr.31\linewidth-1\tabcolsep}}
    \begin{tabular}{MMM}
        \includegraphics[width=6cm,height=4cm]{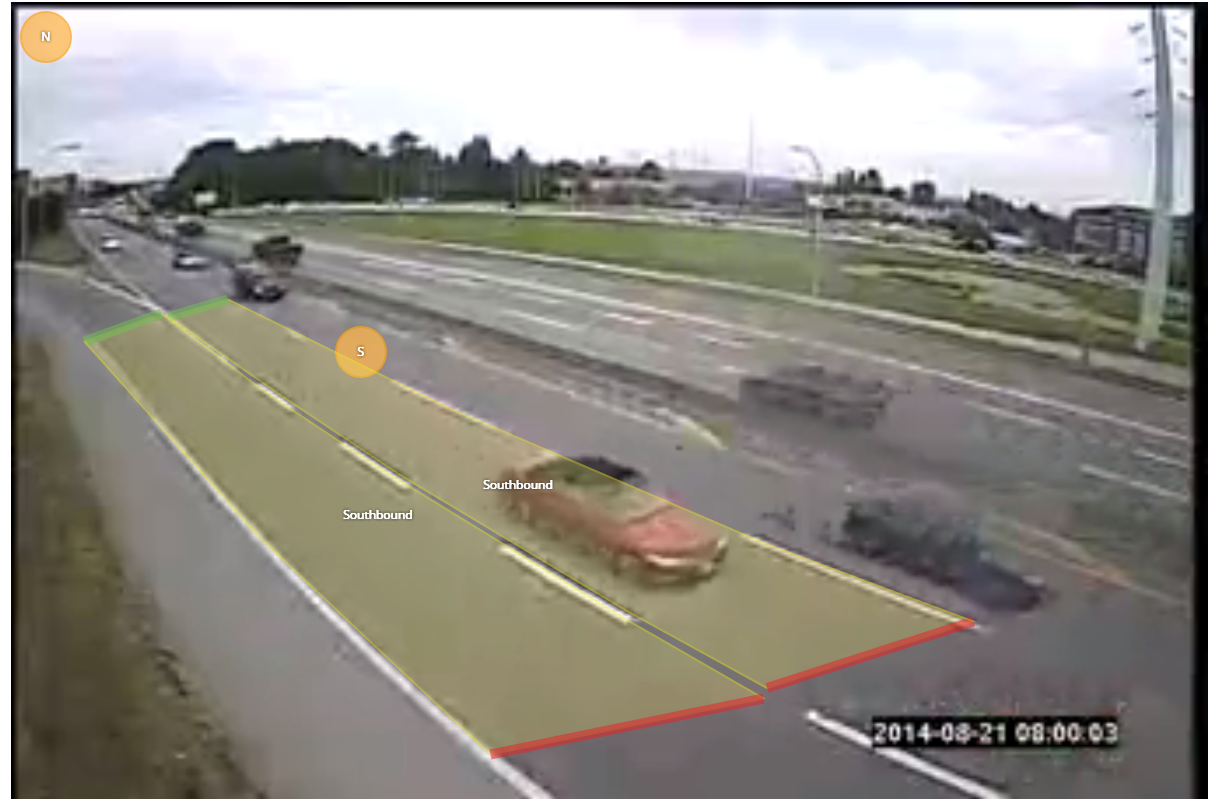} &
        \includegraphics[width=6cm,height=4cm]{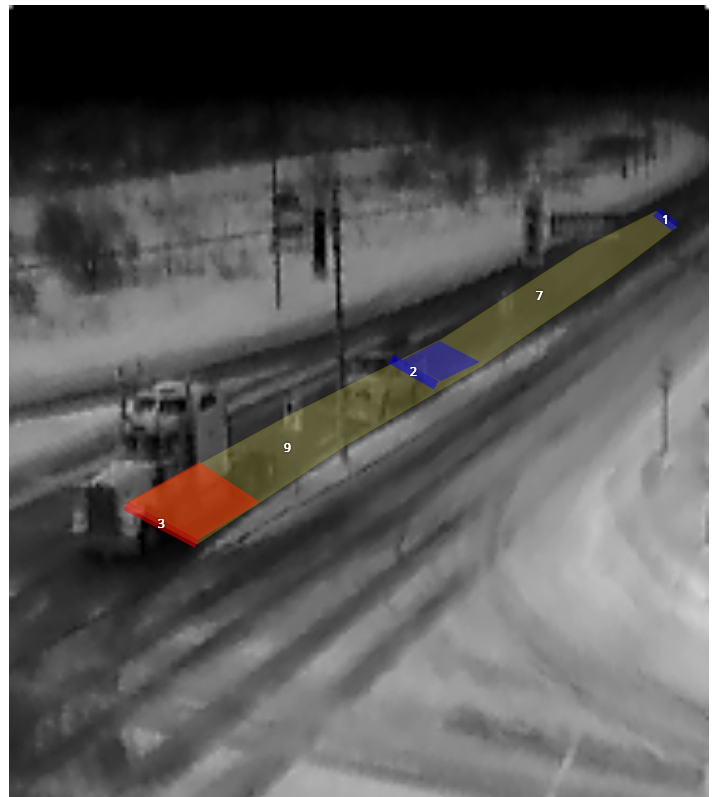} &
        \includegraphics[width=6cm,height=4cm]{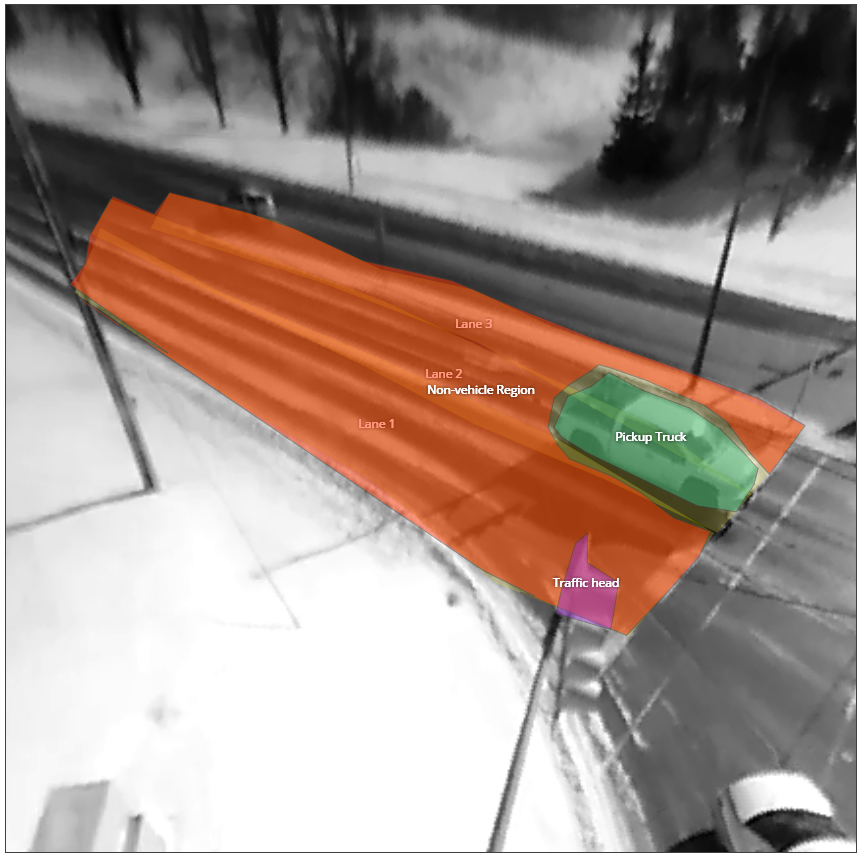} \\
        (a) configuration & (b) localization annotation &(c) boundary
        annotation
    \end{tabular}
    \caption{
    (a) Two southbound vehicle presence detection zones are annotated as yellow polygons. Vehicles enter each zone at the green line segment and exit at the red line.  (b) Three vehicle regions define the lane entry, mid-point and exit. A truck has just reached the exit region and is labelled as a large truck with trailer. (c) Boundaries are drawn around objects of intersect, which are also classified, e.g. road, pick-up truck, or signal head.}
    \label{fig:annotation_gui}
\end{figure*}

 The ITS traffic event dataset currently consists of over $7.7$ million video frames and continues to expand monthly. Each frame can be annotated in several ways. Configuration defines vehicle detection zones relative to the roadway, see~\FigPartref{fig:annotation_gui}{a}. Depending on the study type, this zone may encompass the entire visible lane or a region that is only large enough to fit one vehicle. The annotators also have the ability to label environmental conditions associated with the video or individual frames. The vehicles can be localized allowing the annotators to specify when a vehicle is present at one or more locations, as illustrated in~\FigPartref{fig:annotation_gui}{b} with red and blue regions. The boundaries for objects in the scene can also be drawn by annotators, see~\FigPartref{fig:annotation_gui}{c}. For each case, the annotator can label objects with appropriate classes, e.g. passenger car. Once annotation is complete, a review process ensures that the annotations are correct; a percentage of video frames are annotated by multiple annotators to provide a measure of annotator variability and, similarly, multiple reviews provide reviewer variability metrics. The review process is continuously improved through training to reduce the probability of accepting a misannotated object.

The proposed video dataset is generated using the process detailed in~\Subsecref{subsec:proposed_positive_and_negative_mining}. Initially, a random sample of the acquired video data is obtained. Each video is configured, and for each video frame, every vehicle is localized through human annotation. These localized vehicles become the testing dataset. A subset of video frames are initially randomly sampled from the testing dataset to become the training dataset. Boundaries for each vehicle are annotated for each sample in the training dataset. Using the initial training dataset, the classifier parameters were trained and the results were evaluated against the testing set. Testing samples with poor performance are then given to human annotators for boundary annotation before being moved into the training dataset. Meanwhile, the testing set continues to grow as newly acquired video is randomly sampled and added. The composition of the resulting dataset is detailed in the following sub-section,~\Subsecref{subsec:proposed_dataset}. The proposed dataset contains about $7.7$ million samples of localization and $19,244$ samples of boundary annotations.

\subsection{Proposed large-scale ITS dataset}\label{subsec:proposed_dataset}

The proposed dataset contains $3,082$ five minute video segments, a total of $256.8$ hours, acquired from $15$ locations throughout Canada. Since there are a total of $182$ distinct lanes, each lane is sampled, on average, for about $1.4$ hours, with a corresponding $~16.9$ discontinuous segments of continuous five minute video per lane. The centroid of each vehicle is specified at least once per detection zone. The dataset contains approximately $209$ hours recorded during the day, $121$ hours at night, and $51.3$ hours at dusk or dawn. The dataset contains approximately $4.5$ hours of light rain, $28.4$ hours of heavier rain, $7.5$ hours of light snow, $25.5$ hours of regular snow, and $7.5$ hours of heavy snow. There are also $187.6$ hours of clean roadways, $8$ hours with some snow, $28.9$ hours of snow covered, and $61.2$ of wet roads.

\begin{figure*}
    \centering
    \newcolumntype{M}{>{\centering\arraybackslash}m{\dimexpr.23\linewidth-2\tabcolsep}}
    \newcommand{\colWidth}{4cm}%
    \newcommand{\rowHeightA}{4cm}%
    \newcommand{\rowHeightB}{2cm}%
    \newcommand{\rowHeightC}{3cm}%
    \begin{tabular}{MMMM}
        \begin{overpic}[width=\colWidth,height=\rowHeightA]{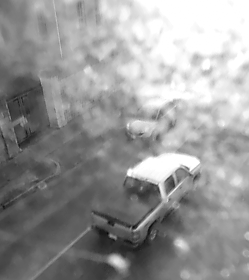}\put(5,5){\textbf{\color{green}(a) glare}}\end{overpic} &
        \begin{overpic}[width=\colWidth,height=\rowHeightA]{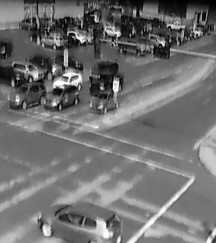}\put(5,5){\textbf{\color{green}(b) partially wet}}\end{overpic} &
        \begin{overpic}[width=\colWidth,height=\rowHeightA]{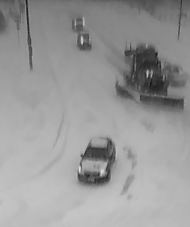}\put(5,5){\textbf{\color{blue}(c) snow}}\end{overpic} &
        \begin{overpic}[width=\colWidth,height=\rowHeightA]{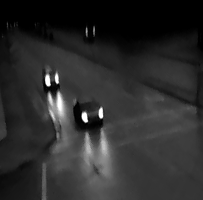}\put(5,5){\textbf{\color{green}(d) reflections}}\end{overpic} \\

        \begin{overpic}[width=\colWidth,height=\rowHeightB]{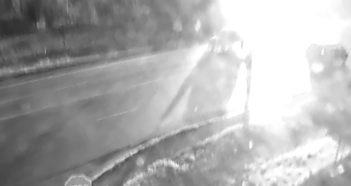}\put(5,5){\textbf{\color{blue}(e) glare}}\end{overpic} &
        \begin{overpic}[width=\colWidth,height=\rowHeightB]{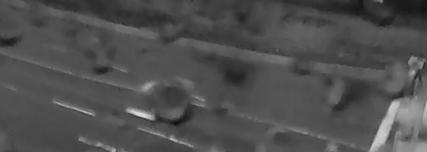}\put(5,5){\textbf{\color{green}(f) rain}}\end{overpic} &
        \begin{overpic}[width=\colWidth,height=\rowHeightB]{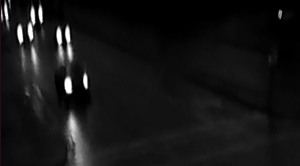}\put(5,5){\textbf{\color{green}(g) reflections}}\end{overpic} &
        \begin{overpic}[width=\colWidth,height=\rowHeightB]{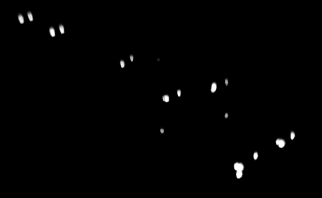}\put(5,5){\textbf{\color{green}(h) night}}\end{overpic} \\

        \begin{overpic}[width=\colWidth,height=\rowHeightC]{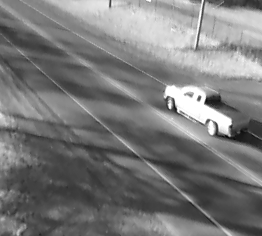}\put(5,5){\textbf{\color{green}(i) shadows}}\end{overpic} &
        \begin{overpic}[width=\colWidth,height=\rowHeightC]{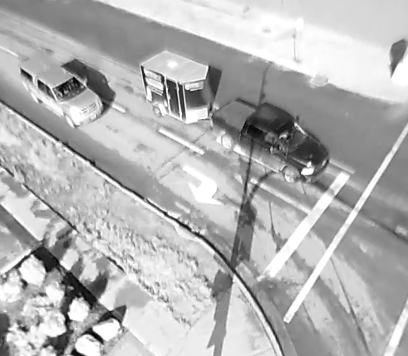}\put(5,5){\textbf{\color{blue}(j) partially wet}}\end{overpic} &
        \begin{overpic}[width=\colWidth,height=\rowHeightC]{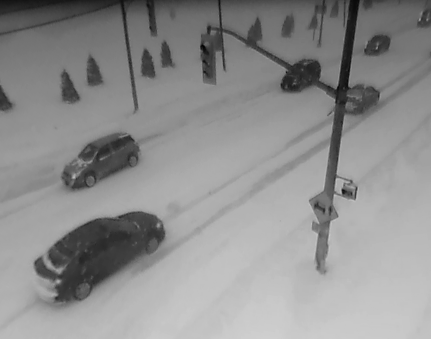}\put(5,5){\textbf{\color{blue}(k) snow}}\end{overpic} &
        \begin{overpic}[width=\colWidth,height=\rowHeightC]{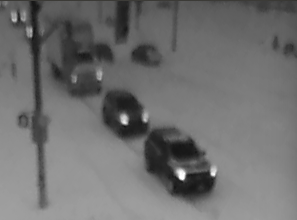}\put(5,5){\textbf{\color{blue}(l) snow}}\end{overpic}
    \end{tabular}
    \caption{These figures represent several of many conditions, resolutions, camera perspectives, and locations contained in the proposed ITS dataset.}
    \label{fig:dataset_examples}
\end{figure*}

\begin{figure*}
    \centering
    \begin{tabular}{cc}
        \includegraphics[width=0.5\textwidth]{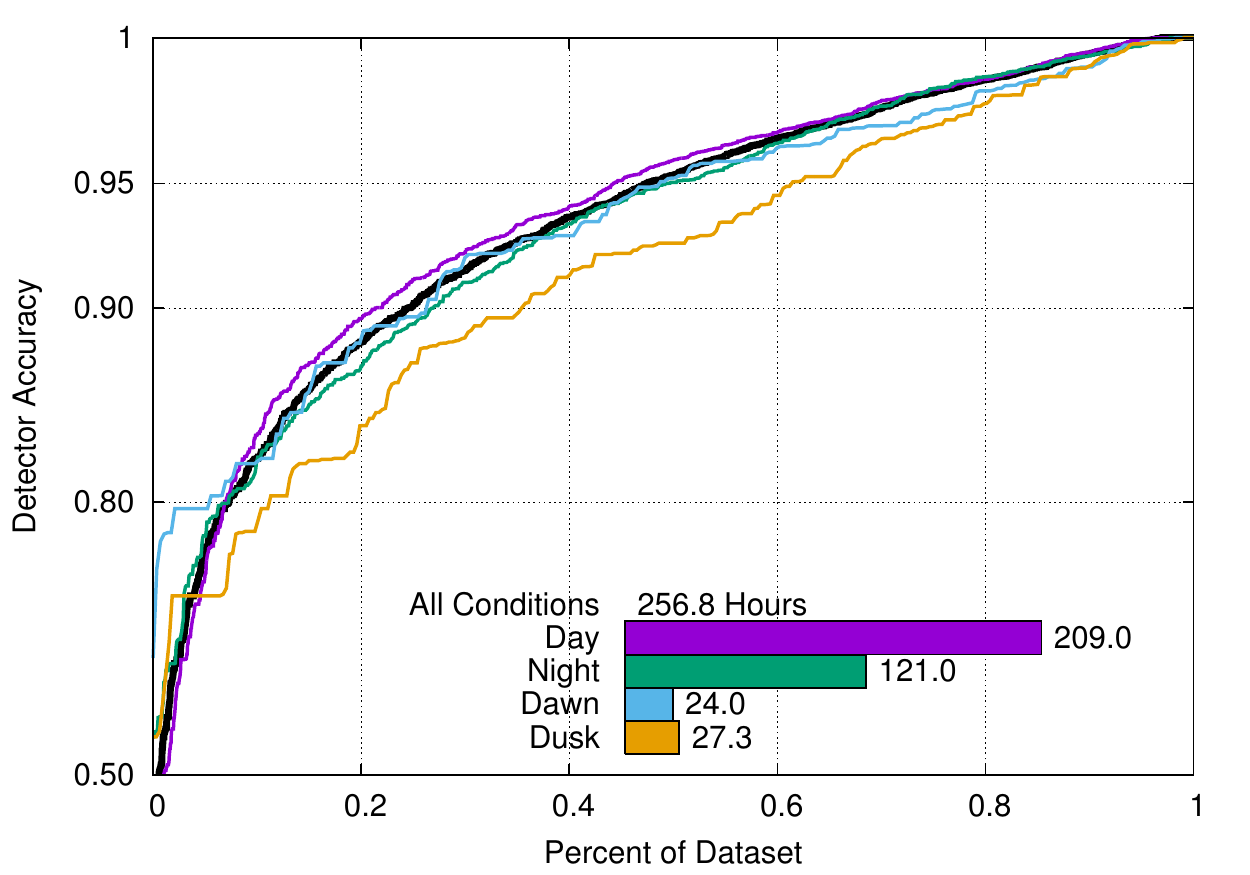} & \includegraphics[width=0.5\textwidth]{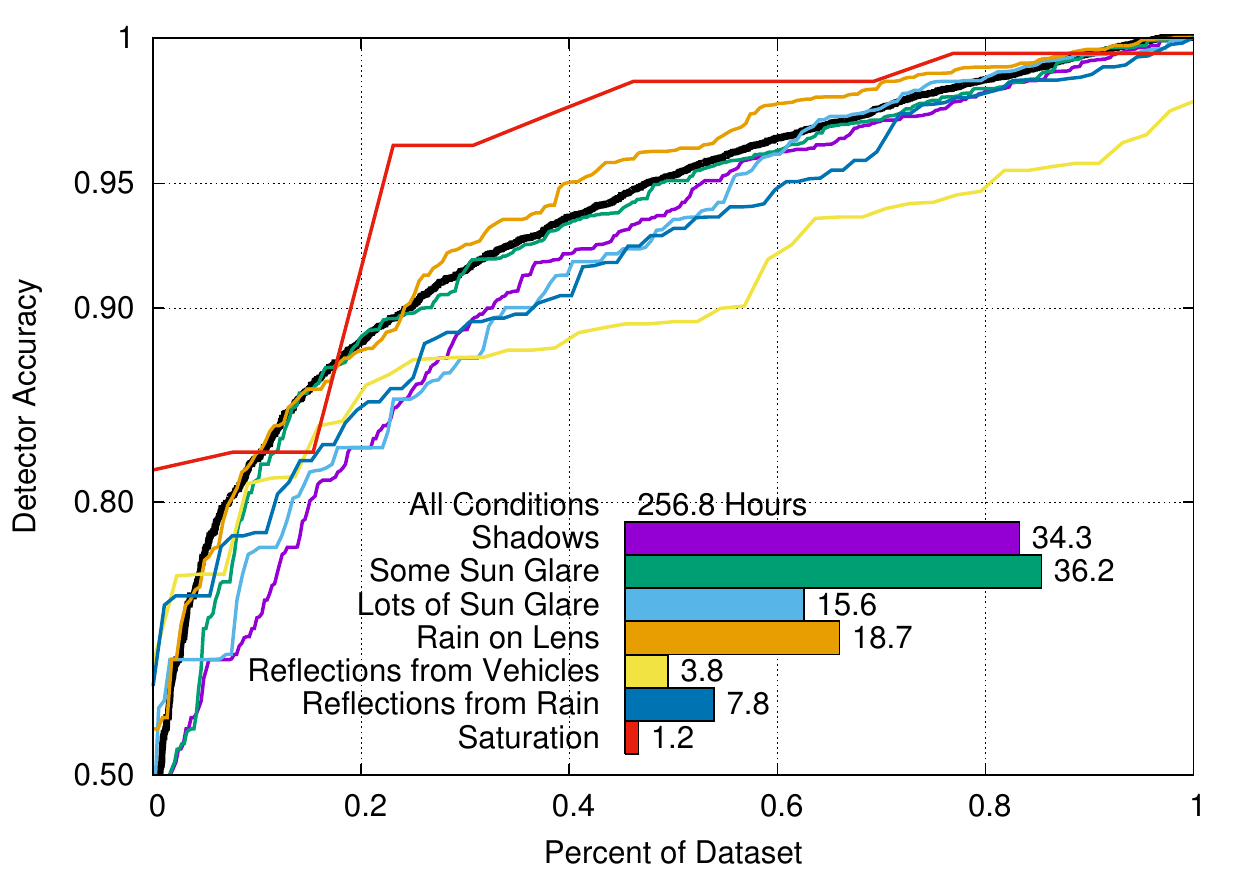} \\
        (a) time of day & (b) camera properties \\
        \\
        \includegraphics[width=0.5\textwidth]{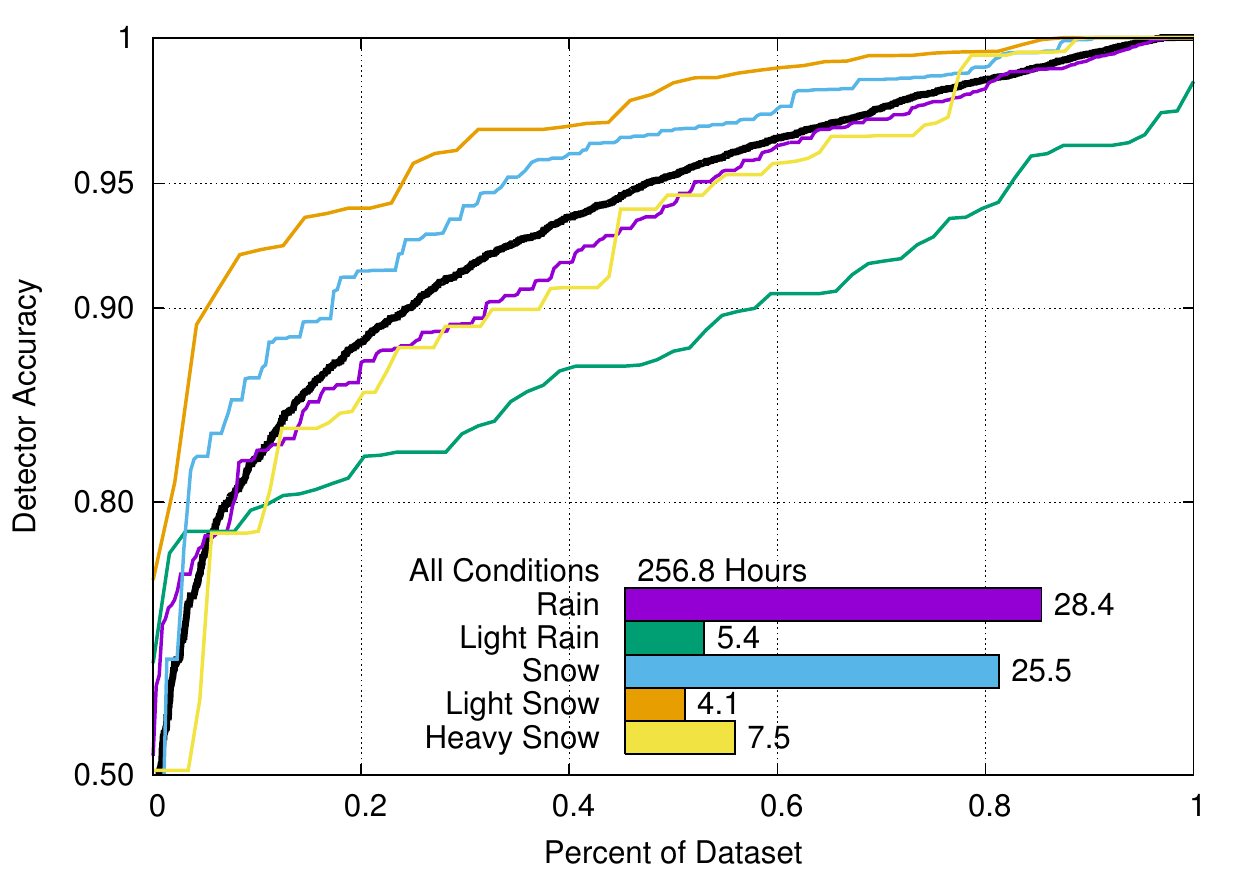} & \includegraphics[width=0.5\textwidth]{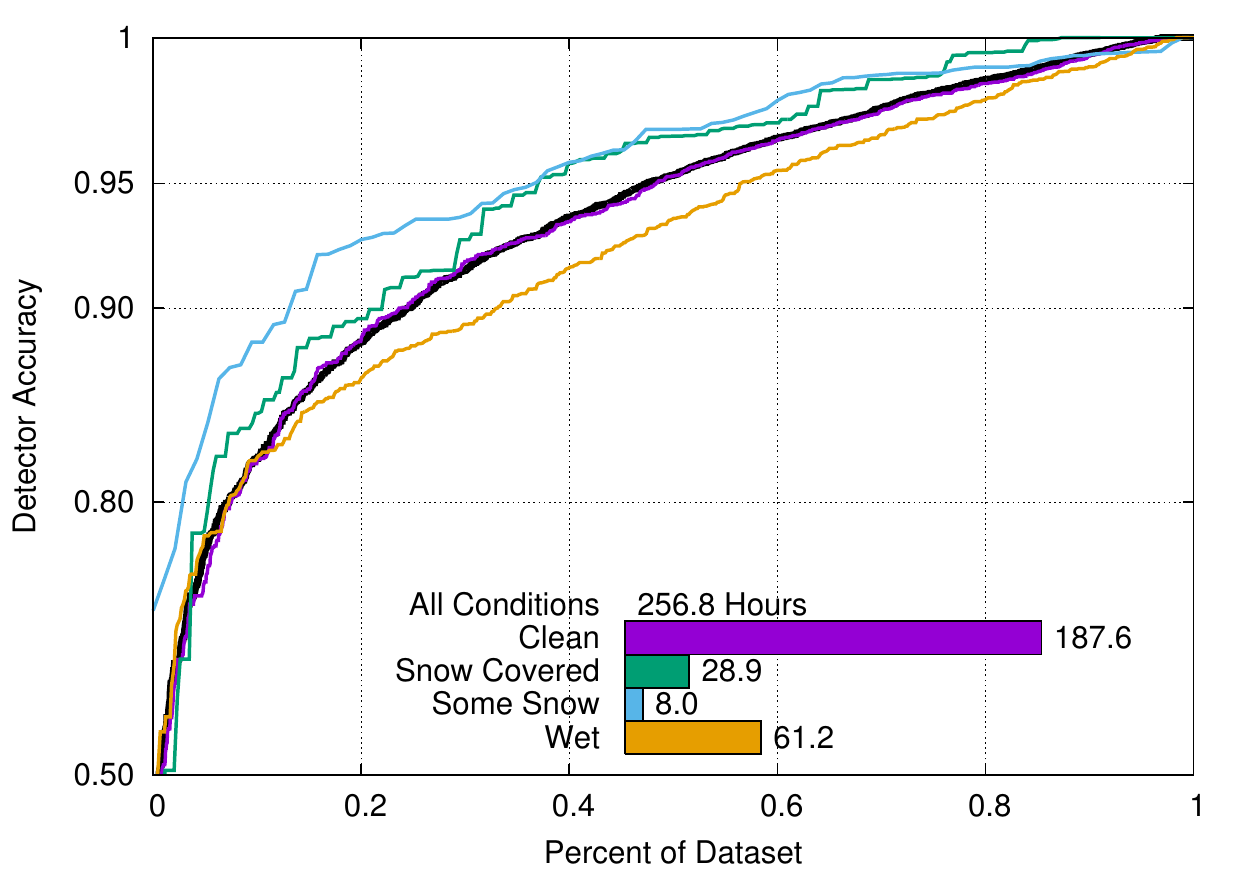} \\
        (c) precipitation types & (d) road conditions \\
    \end{tabular}
    \caption{Cumulative distributions of four annotation categories. From time of day (a), the trained video detector performs better at dawn than at dusk (27.3 hours of video). The authors' video detector detects vehicles through headlights during the night, and performs better at dawn because it was observed that drivers typically have headlights on at dawn more often than they have headlights on at dusk. From observing (b) camera properties and (c) precipitation types, the detector performs worse in light rain conditions and also when reflections from vehicles are present. Also from observing (c) precipitation types and (d) road conditions, it should be noted that the video detector performs very well in snow and snow covered conditions, which may be due to the higher contrast between vehicles and their surrounding.}
    \label{fig:dataset_composition_vs_accuracy}
\end{figure*}

\Figref{fig:dataset_examples} illustrates several examples of various conditions represented in the dataset. \Secref{sec:experimental_validation} presents detailed breakdown of detector accuracy when evaluated against additional sets of conditions,~\Figref{fig:dataset_composition_vs_accuracy}. Each video frame contains vehicles with resolution between $14 \times 14$ and about $80 \times 80$ pixels, with median vehicles resolution as $26 \times 26$ pixels. The video data has been acquired over a span of $2$ years, with a range of weather and operating conditions.

\subsection{Vehicle event detection system}
As described in a previous work~\cite{MiovisionCVPR2014}, the trained classifier parameters are incorporated into a vehicle event detection system, which utilizes background modelling, time of day estimation, object detection, Kalman filter based tracking, and AdaBoost. Each intersection is manually configured with view specific metadata including a collection of incoming and outgoing lanes or zones. The vehicle classifier is combined with background subtraction and an online updated parametric lane model to associate vehicles to lanes and zones for vehicle presence detection. The classifier parameters are trained using the proposed ITS dataset and continuous learning process.

\subsection{Evaluation metrics}\label{subsubsec:evaluation_metrics}
Video detector accuracy is reported based on the four detector modes defined in the NEMA TS-2 standard~\cite{NEMATS2}. \begin{enumerate}
    \item Pulse: a pulse of duration $100$ to $150~ms$ that is triggered when a vehicle enters the detection zone.
    \item Controlled output: identical to pulse, but with a configurable pulse duration.
    \item Continuous presence: a signal is generated for as long as a vehicle is present in the detection zone.
    \item Limited presence: identical to continuous presence, but with a configurable maximum duration. Note that the signal may end before the maximum duration if the detected vehicle leaves the detection zone early
\end{enumerate} Throughout this paper, presence mode is used exclusively because the authors have focused on detection applications requiring this metric. Other applications and systems require some or all of these operational modes.

The evaluation system reports a confusion matrix representing the following items. \begin{enumerate}
    \item True positive, $\TP$, (true call): a vehicle is present and a corresponding detection call is correct.
    \item False negative, $\FN$, (false call): a vehicle is not present in the detection zone, despite a video detection. Detectors may fail in this way if a vehicle was previously in the detection zone, but did not detect the vehicle leaving the zone, also known as a `stuck on call'\cite{medina2008vol1}.
    \item False positive, $\FP$, (missed call): a vehicle is present in the detection zone, but is not detected. This failure may occur if the detector initially identifies the vehicle, but fails to continuously detect the vehicle the entire time it is in the zone, also known as a `dropped call'\cite{medina2008vol1}.
    \item True negative, $\TN$, (true non-call) a vehicle is not present in the detection zone, which the detector correctly reports.
\end{enumerate} The overall accuracy, $\a$, is derived from the confusion matrix, \begin{align}
    \a &= \Biggl( \frac{\TP+\TN}{\TP+\FN+\FP+\TN} \Biggr).
    \label{eq:overall_accuracy}
\end{align}

In addition to the metrics above, the evaluation system also measures runtime, and receiver operating characteristic (ROC) curves related to customer accepted ratio of true positives compared to false positives. Runtime is a useful metric particularly if a real-time detection algorithm is being evaluated.

\subsection{Overall evaluation accuracy}
The overall accuracy is evaluated on the testing dataset described in~\Subsecref{subsec:proposed_dataset}. The vehicle detection accuracy for each video is sorted from lowest to highest is $1/2$ of the dataset have a vehicle presence accuracy in excess of $95\%$ and that $19/20$ of the dataset have an accuracy in excess of $78\%$. In addition, for $1/2$ of the dataset, counting metrics exceed $78\%$ accuracy.

\subsection{Dataset composition vs. accuracy}
\Figref{fig:dataset_composition_vs_accuracy} illustrates the accuracy of the authors' video based vehicle detector using the proposed ITS dataset for training and testing. The overall detection accuracy is illustrated, the solid black curve exceeds $95\%$ for $1/2$ and $78\%$ for $19/20$ of the dataset. Using the annotation labels, accuracy for the video detector in a variety of conditions can be calculated. The four sub-figures in~\Figref{fig:dataset_composition_vs_accuracy} present comparisons of time of day, environmental conditions, precipitation types, and road conditions. From these figures, researchers can identify scenarios with the lowest accuracy and improve the detection accuracy by modifying and evaluating proposed video detector designs and by improving the training process by incorporating more training samples into the dataset for poorly performing scenarios. There are additional annotation labels that are collected but not illustrated below, such as traffic density, distance of the lane from the camera, number of adjacent incoming or outgoing lanes, and distributions of vehicle classes.

\subsection{Sample size vs. evaluation accuracy}
\begin{figure}
    \centering
    \includegraphics[width=9cm]{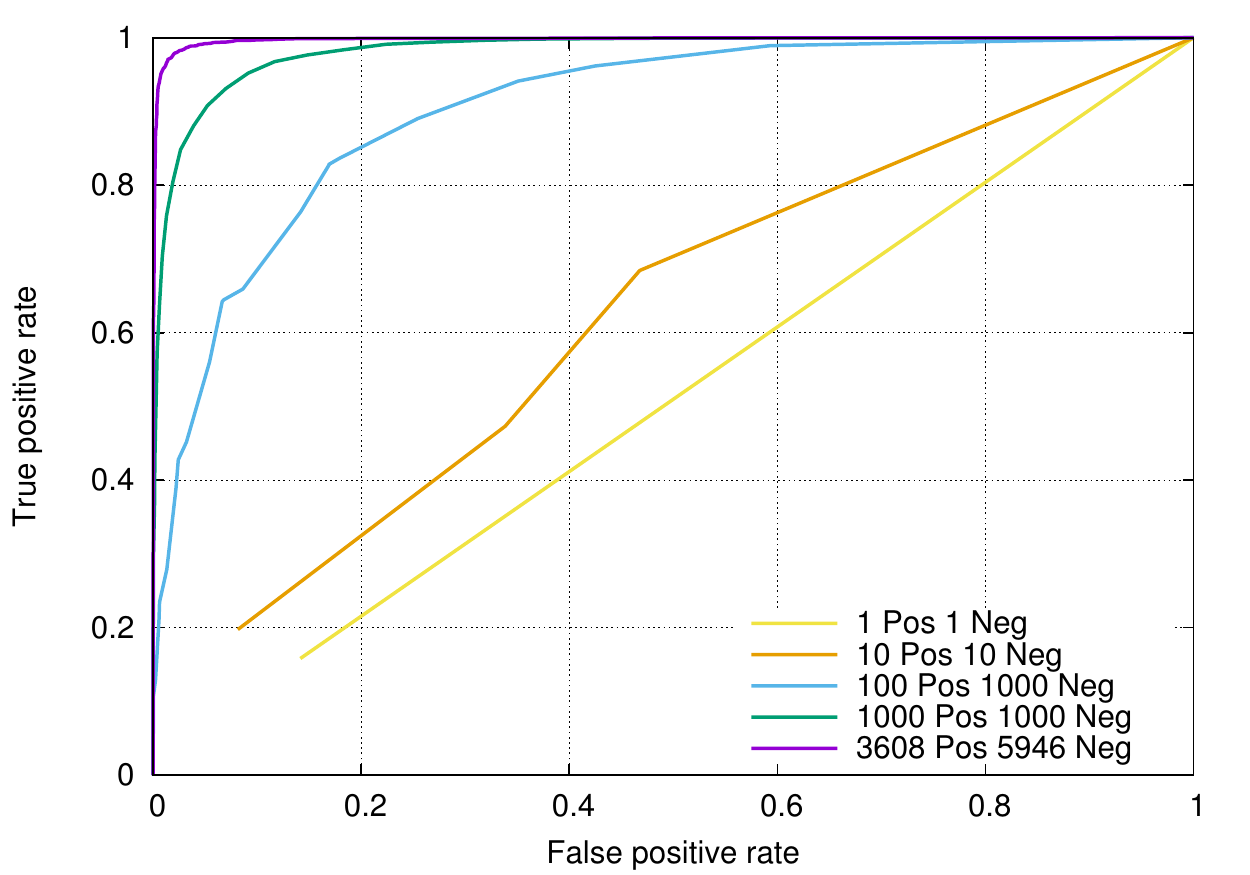}
    \caption{ROC curve illustrating that the detection classifier achieves the maximum true positives and minimum false positives.}
    \label{fig:roc_varying_sample_sizes}
\end{figure}

Sample quantity and diversity play a significant role in  designing a real-time and continuous learning based vehicle detection system. A positive and negative mining technique is applied to collect a diverse training dataset. The receiver operating characteristic (ROC) curve, illustrated in \Figref{fig:roc_varying_sample_sizes}, indicates that the detector accuracy, evaluated on the testing set, increases as the quantity of training samples increases. The primary reason for this phenomenon is that a reasonable quantity of training samples is required to establish sufficient diversity to represent the testing dataset. Further, \Figref{fig:roc_varying_sample_sizes} indicates that the  designed AdaBoost  based strong classifier uses a threshold of $0.058301$, similar to the ideal value of $0$, to maximizes true positives and minimizes false positives when deployed in field.

\section{Discussion and conclusions}\label{sec:conclusion}
This paper presents an ITS dataset for the purpose of real-time vehicle detection. The proposed positive and negative mining process allows the creation of an ITS dataset by selecting training and testing samples that are representative of the real-world. The process also culls the dataset by removing noisy and redundant samples. As shown, a detector can be trained using significantly fewer, but representative, samples than using an entire dataset. Positive and negative mining avoided the need to annotate~$7.7$ million video frames containing vehicle boundaries and allowed the detector to be trained on only $19,244$ boundaries instead. The detector achieved $95\%$ accuracy for $1/2$ and over $78\%$ accuracy for $19/20$ of the data, evaluated on all $256.8$ hours of the video, containing~$7.7$ million video frames of localized vehicles; localization is significantly more efficient than boundary annotation.

This shows the effectiveness of continuous learning applied to real-time vehicle detection. The continuous learning process utilizes positive and negative mining to efficiently represent a diverse and compact training dataset from a massive large-scale population of roadways and operating environments. The process has generated, to the best of the authors' knowledge, the largest and most diverse ITS dataset to date. Further, the continuously trained detector parameters, using a large-scale distributed computing system, are transmitted and incorporated into real-world video detectors as the ITS dataset continues to update. The learning process and training system allow researchers to quickly evaluate new algorithms, not only for detection, but for future ITS applications; the researcher can focus on algorithm design instead of data management.

The work presented here is only the beginning. The authors plan to expand the scope of the ITS dataset by including data acquired from all over the world and to start annotating additional ITS classes, such as pedestrians, bicycles, buses, and various classes of trucks. The authors also fully intend to provide API access to this dataset, allowing the computer vision researcher community to evaluate and train their own algorithms on a large-scale. As for the demonstrated video detector, the immediate goal is to improve accuracy for rainy and highly reflective conditions.

\ifCLASSOPTIONcaptionsoff
  \newpage
\fi

\bibliographystyle{IEEEtran}
\bibliography{IEEEabrv,short,large_scale_learning_and_evaluation}

\newcommand{\photoW}{1in}%
\newcommand{\photoH}{1.5in}%

\pagebreak
\begin{IEEEbiography}[{\includegraphics[width=\photoW{},height=\photoH{},clip,keepaspectratio]{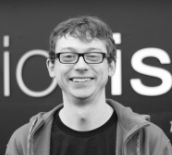}}]{Justin A. Eichel}
computer vision architect at Miovision Technologies. PhD from Vision and Image Processing Lab, Systems Design Engineering, University of Waterloo. Previous experience as self-employed consultant related to multi-spectrum tracking and medical image processing researcher. Justin is skilled at statistical modelling, pattern recognition, and machine learning.
\end{IEEEbiography}

\begin{IEEEbiography}[{\includegraphics[width=\photoW{},height=\photoH{},clip,keepaspectratio]{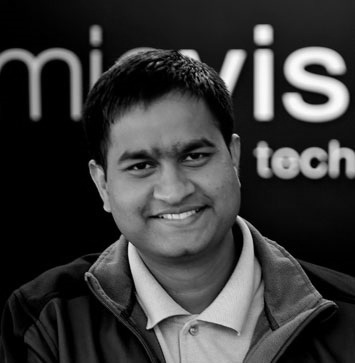}}]{Akshaya Mishra}
applied research scientist at Miovision Technologies and concurrently holds an Adjunct faculty position at the University of Waterloo. He holds a PhD from the University of Waterloo, as well as an M. Tech and B.E. in Electrical Engineering from the Indian Institute of Technology. Akshaya focuses in statistical modelling, image processing, and pattern recognition.
\end{IEEEbiography}

\begin{IEEEbiography}[{\includegraphics[width=\photoW{},height=\photoH{},clip,keepaspectratio]{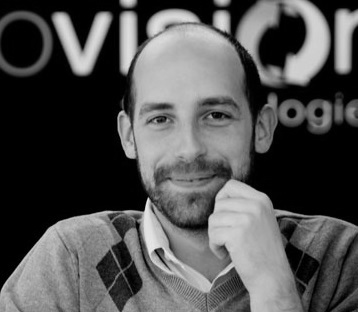}}]{Nicholas Miller}
applied research scientist at Miovision Technologies having received his master's and bachelor's of Mathematics from the School of Computer Science at the University of Waterloo. He has studied visual perception of events and physical interactions and is experienced in video sensing, distributed computing systems, machine learning, and optimization.
\end{IEEEbiography}

\begin{IEEEbiography}[{\includegraphics[width=\photoW{},height=\photoH{},clip,keepaspectratio]{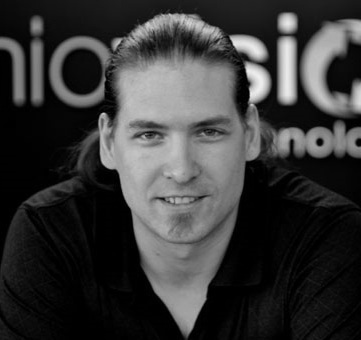}}]{Nicholas Jankovic}
embedded developer at Miovision Technologies and is a licensed Professional Engineer of Ontario and holds a Master of Engineering Science in Sensing and Mechatronic Systems from the University of Western Ontario. He is experienced in machine vision, has designed camera systems, and embedded software development and verification.
\end{IEEEbiography}

\begin{IEEEbiography}[{\includegraphics[width=\photoW{},height=\photoH{},clip,keepaspectratio]{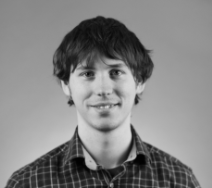}}]{Tyler Abbott}
software developer at Miovision Technologies and has received an Ontario College Advanced Diploma in Computer Science Technology with High Honours from Sheridan College. Tyler has previously worked designing software at Blackberry and has experience with software development, rapid prototyping, and data visualization.
\end{IEEEbiography}

\begin{IEEEbiography}[{\includegraphics[width=\photoW{},height=\photoH{},clip,keepaspectratio]{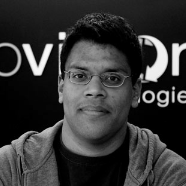}}]{Mohan A. Thomas}
software developer at Miovision Technologies and has obtained his Bachelor’s of Applied Science from Systems Design Engineering at the University of Waterloo. Mohan is experienced with traffic engineering and systems integration, and has previous experience at Nuance Communications Inc. and National Research Council Canada.
\end{IEEEbiography}

\begin{IEEEbiography}[{\includegraphics[width=\photoW{},height=\photoH{},clip,keepaspectratio]{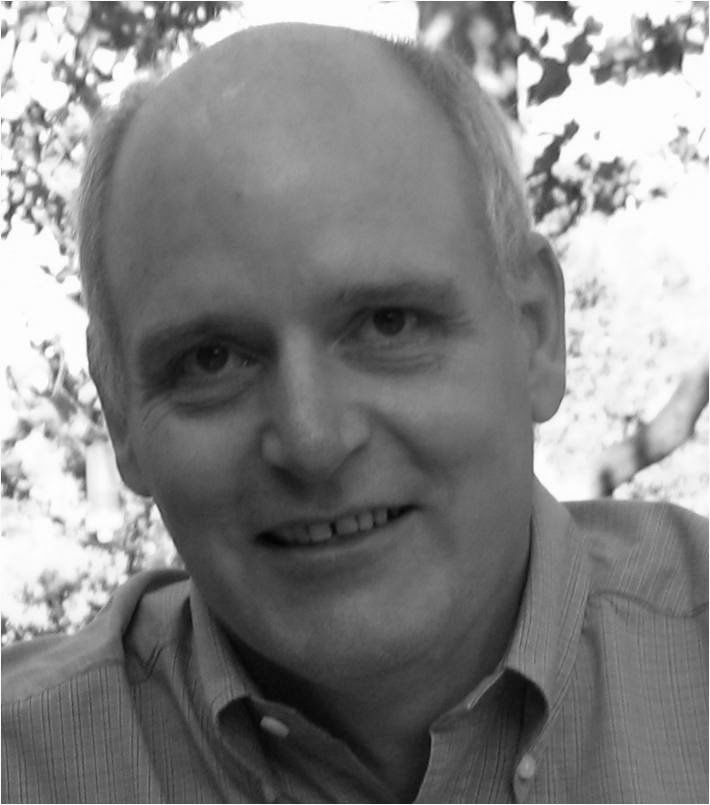}}]{Doug Swanson}
VP Engineering at Miovision Technologies. Responsible for the MioLabs Research team developing computer vision, traffic simulation and optimization technologies and has previous experience at Blackberry, Cisco and Nortel. Doug obtained a Bachelor of Applied Science in System Design Engineering from the University of Waterloo.
\end{IEEEbiography}

\begin{IEEEbiography}[{\includegraphics[width=\photoW{},height=\photoH{},clip,keepaspectratio]{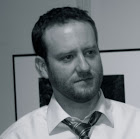}}]{Joel Keller}
an embedded software specialist at Miovision Technologies,
experienced in firmware development, hardware/software co-design, and software architecture. Joel holds a Bachelor's of Mathematics in Computer Science from the University of Waterloo.
\end{IEEEbiography}
\end{document}